\documentclass[preprint,12pt]{elsarticle}
\usepackage[margin=1.8 cm]{geometry}
\usepackage{amssymb}
\usepackage{amsmath}
\usepackage{amsthm}
\usepackage{natbib}
\usepackage{graphicx}
\usepackage{float}
\usepackage{hyperref}
\usepackage{cleveref}
\usepackage{caption}
\usepackage{multicol}
\usepackage{soul}
\usepackage{tikz,pgfplots}
\usepackage{dblfloatfix}
\usepackage{textgreek}
\usepackage{tabulary}
\usepackage{ragged2e}
\usepackage{subcaption}
\usepackage{bm}
\journal{Engineering Applications of Artificial Intelligence}

\begin{document}

\begin{frontmatter}

%% Title, authors and addresses

%% use the tnoteref command within \title for footnotes;
%% use the tnotetext command for theassociated footnote;
%% use the fnref command within \author or \affiliation for footnotes;
%% use the fntext command for theassociated footnote;
%% use the corref command within \author for corresponding author footnotes;
%% use the cortext command for theassociated footnote;
%% use the ead command for the email address,
%% and the form \ead[url] for the home page:
%% \title{Title\tnoteref{label1}}
%% \tnotetext[label1]{}
%% \author{Name\corref{cor1}\fnref{label2}}
%% \ead{email address}
%% \ead[url]{home page}
%% \fntext[label2]{}
%% \cortext[cor1]{}
%% \affiliation{organization={},
%%             addressline={},
%%             city={},
%%             postcode={},
%%             state={},
%%             country={}}
%% \fntext[label3]{}

\title{Multi-Task Lane-Free Driving Strategy for Connected and Automated Vehicles: A Multi-Agent Deep Reinforcement Learning Approach}

%% use optional labels to link authors explicitly to addresses:
\author{Mehran Berahman\corref{cor1}\fnref{label1}}
\cortext[cor1]{Corresponding author}
\ead{mnberahman@hafez.shirazu.ac.ir}
\affiliation[label1]{organization={Department of Electrical and Computer Engineering, Shiraz University},city={Shiraz},country={Iran}}

\author[label2]{Majid Rostami-Shahrbabaki}
\ead{majid.rostami@tum.de}
\author[label2]{Klaus Bogenberger}
\affiliation[label2]{organization={Chair of Traffic Engineering and Control, Technical University of Munich},city={Munich},
            country={Germany}}           
\ead{klaus.bogenberger@tum.de}
%% Abstract
\begin{abstract}
Deep reinforcement learning has shown promise in various engineering applications, including vehicular traffic control. The non-stationary nature of traffic, especially in the lane-free environment with more degrees of freedom in vehicle behaviors, poses challenges for decision-making since a wrong action might lead to a catastrophic failure. In this paper, we propose a novel driving strategy for Connected and Automated Vehicles (CAVs) based on a competitive Multi-Agent Deep Deterministic Policy Gradient approach. The developed multi-agent deep reinforcement learning algorithm creates a dynamic and non-stationary scenario, mirroring real-world traffic complexities and making trained agents more robust. The algorithm's reward function is strategically and uniquely formulated to cover multiple vehicle control tasks, including maintaining desired speeds, overtaking, collision avoidance, and merging and diverging maneuvers.  Moreover, additional considerations for both lateral and longitudinal passenger comfort and safety criteria are taken into account. We employed inter-vehicle forces, known as nudging and repulsive forces, to manage the maneuvers of CAVs in a lane-free traffic environment. The proposed driving algorithm is trained and evaluated on lane-free roads using the Simulation of Urban Mobility platform. Experimental results demonstrate the algorithm's efficacy in handling different objectives, highlighting its potential to enhance safety and efficiency in autonomous driving within lane-free traffic environments.
\vspace{10pt}
\end{abstract}

%% Keywords
\begin{keyword}
%% keywords here, in the form: keyword \sep keyword
Multi-agent deep deterministic policy gradient\sep reinforcement learning\sep lane-free traffic\sep connected and automated vehicles\sep traffic control.
%% PACS codes here, in the form: \PACS code \sep code

%% MSC codes here, in the form: \MSC code \sep code
%% or \MSC[2008] code \sep code (2000 is the default)

\end{keyword}

\end{frontmatter}

%% Add \usepackage{lineno} before \begin{document} and uncomment 
%% following line to enable line numbers
%%\linenumbers

%% main text
%%

%% Use \section commands to start a section

\section{Introduction}\label{sec_1}
Connected and automated vehicles (CAVs) have garnered significant interest from both the academic and industrial sectors \citep{shladover2018connected, schwarting2018planning}. CAVs are equipped with advanced, high-precision sensors and intelligent technologies that facilitate swift and reliable communication. This communication extends to various aspects, including interactions between vehicles and infrastructure using vehicle-to-vehicle  (V2V) and vehicle-to-infrastructure (V2I) communication, respectively \cite{9856630}. At a high level of automation, they can perform various driving tasks, including acceleration, braking, steering, and lane changing, without the need for human involvement. These merits of CAVs bring several public advantages, including increased safety on the roads, leading to fewer accidents, reduced traffic congestion, and lower emissions \citep{spielberg2019neural}.\par 
 Driving is a multifaceted activity with intricate interactions that cannot be comprehensively articulated through predetermined rules. Hence, autonomous systems must not solely depend on predefined rules to address every potential road situation. Instead, approaches as adaptive learning agents can be implemented that have the capacity to enhance their driving proficiency by continuously refining their skills through learning from real-world or simulated experiences and exploration. The deep reinforcement learning (DRL) framework \citep{wang2020deep, kiran2021deep}  provides a flexible learning-based platform for creating adaptive solutions to address such issues. Therefore, DRL applied to the decision-making processes of autonomous vehicles has become a focal point in recent research and studies \citep{kiran2021deep, wurman2022outracing}.\par
 In this work, we put forward a DRL-based coordinated decision-making method mainly implemented on the lane-free traffic (LFT) paradigm. In 2021, Papageorgiou et al. \citep{papageorgiou2021lane} introduced an innovative approach to freeway traffic in the fully automated and connected environment where the need for conventional lanes is questioned. This concept allows vehicles to utilize the entire road width instead of being confined to specific fixed lanes. This novel concept can significantly enhance traffic management efficiency, as recent assessments have shown that a fully automated lane-free environment could greatly boost road capacity \citep{berahman2022driving,malekzadeh2022empirical, yanumula2023optimal}.\par
The domain of vehicular traffic is marked as a non-stationary environment, particularly in a lane-free context where vehicle behaviors could exhibit notable status change in both lateral and longitudinal directions; therefore, in this emerging field of research, there seems to be a long road ahead to develop reliable driving strategies that can be adeptly applied to these dynamic traffic conditions. In light of this fact, autonomous driving challenges could encompass unforeseeable shifts in the environment and the possibility of unavoidable perception errors, potentially leading an autonomous vehicle down an unsafe path and possibly resulting in severe traffic accidents. Given the gravity of these potential risks, it is imperative to prioritize the development of decision-making systems that can withstand and adapt to environmental uncertainties with a high degree of resilience in the deployment of CAVs' driving strategies \citep{DERUYTTERE2021104257}.\par
One approach to tackling the perception error in a non-stationary environment is employing multi-agent deep reinforcement learning (MADRL) approaches \citep{CHOL2023106229}. Traffic situations are inherently characterized by partial observability, meaning that each vehicle, as an individual agent, does not have a complete view of the entire environment. MADRL offers a promising avenue to enhance collective knowledge by fusing partial observations from several agents, thereby expanding the overall information available for decision-making \citep{schmidt2022introduction}. Moreover, unlike single-agent learning frameworks, where it's common to overlook the influence of other agents in the environment or even to disregard their presence altogether, MADRL learning frameworks offer the capability to explicitly account for and model the interactions among other agents in the environment. \par
To the best of our knowledge, limited attention has been given to developing driving strategies for CAVs in a lane-free environment with MARL. Specifically considering comfort and safety issues for realistic traffic networks that involve complex merging and diverging maneuvers at on-ramp and off-ramps, respectively. In view of the mentioned challenges and the research gap, this work's primary contributions lie in:
\begin{itemize}
\item A multi-agent deep deterministic policy gradient (MADDPG) algorithm \citep{lowe2017multi} is implemented to train the agents of CAVs in a lane-free environment, accounting for the non-stationary nature of this particular traffic system. We construct a MADRL algorithm in an unknown and complex environment where each agent cannot fully perceive the complete information of its environment. In this regard, the decision process fits the framework of partially observable Markov decision processes (POMDPs)  \citep{lowe2017multi}; centralized training and decentralized execution MADRL algorithm is implemented to solve this problem.
\item A novel approach is employed to calculate the nudging and repulsion forces different from the initial definition introduced by the concept of LFT \citep{papageorgiou2021lane}. To this end, a dynamic elliptical safety zone is defined around each vehicle, i.e., each agent. These forces serve as a means to prevent collisions and facilitate overtaking maneuvers among CAVs, enhancing the safety and fluidity of traffic flow. In addition, we mimic the conventional overtaking process by proposing the nudging force to guide vehicles to the right and repulsive force leading them to the left side of the road. This innovation suggests a more structure lane-free traffic with fast-moving vehicles driving on the left.
\item A comprehensive reward function is designed in such a way as to cover multiple objectives of the vehicles (agents). The primary objectives of the agents reflected in the reward function revolve around pursuing their desired speeds and collision avoidance while ensuring passenger comfort and safety criteria.  Thus, fulfilling the aforementioned objectives in trained agents equips them with the capability to perform multitasking behavior. Importantly, this strategy is also extended to encompass smooth merge and diverge maneuvers by imposing virtual nudge and repulsion forces on CAVs in specific circumstances.
\end{itemize}
\par It is worth mentioning that the training and initial evaluation of the developed MADRL algorithm is carried out on a lane-free ring road, and the assessment of the trained agents is evaluated on a 4-$km$ freeway stretch with on- and off-ramps at specific locations. To this end, we customized the Simulation of Urban Mobility (SUMO) platform \citep{behrisch2011sumo, lopez2018microscopic}  to accommodate the specific requirements for simulating lane-free traffic scenarios that are not inherently supported by SUMO.\par
The structure of the rest of this paper unfolds as follows: The related works are documented in Section \ref{sec_2}. In Section \ref{sec_3}, we introduce the preliminary concepts. Section \ref{sec_4} provides a detailed presentation of the proposed MADRL framework. The experimental results are scrutinized in Section \ref{sec_5}. Lastly, Section \ref{sec_6} concludes the paper.

\section{Related works}\label{sec_2}
Several research has been conducted on the concept of lane-free traffic. In \citep{sekeran2022lane}, an exploration of current developments in LFT research and required technologies is provided. The paper offers an overview of existing methods for implementing LFT and delves into the necessary policy considerations for the seamless integration of LFT into the current traffic infrastructure. There are several driving strategies developed for CAVs in freeway lane-free environments, including optimal control approach \citep{yanumula2023optimal,LEVY2022103813}, potential line strategy \citep{Potentialline}, and ad-hoc vehicle movement strategy \citep{malekzadeh2022empirical}. The LFT concept has also been employed in urban networks, for example, in a large roundabout \citep{naderi2022automated} or urban intersection \citep{amouzadi2022optimal}, and integration of vulnerable road users \citep{stueger2023integrated}. Moreover, LFT opens a lot of research opportunities to be explored, including vehicle flocking \citep{rostami2023modeling, rostami2022two}.\par
DRL has found applications in a wide range of intricate tasks related to controlling CAVs, both in lane-based and lane-free traffic environments.
In lane-based traffic, deployment of DRL approaches can be broadly categorized into three main groups: longitudinal control, i.e., longitudinal speed control using acceleration and deceleration manipulation, \citep{liao2020decision, zhang2018human, futuretransp2040057}, lateral control, i.e., lane keep assist, lane changing maneuvers and target lane selection, \citep{sallab2017deep, dong2021space, wang2021harmonious, mirchevska2018high}, and coordinated decision-making (manipulation of both lateral and longitudinal maneuvers of CAVs simultaneously) including features of collision avoidance and merging  \citep{wang2021decision, chen2020deep,saxena2020driving, hoel2018automated, zhang2020adaptive}.\par
Reinforcement learning has also been used to develop driving strategies for CAVs in LFT. In an earlier work, the authors developed an integrated control strategy combining a cruise controller and the deep deterministic policy gradient (DDPG) algorithm applied to manage CAVs in a lane-free environment \citep{berahman2022driving}. This strategy incorporates artificial forces to proactively avert collisions, ensuring safety in both lateral and longitudinal directions. A longitudinal and lateral coordinated decision-making method with a DDPG algorithm is also advanced in \citep{karalakou2023deep}. In \citep{troullinos2021collaborative}, a MADRL approach is employed for freeway driving assistance systems. This method leverages the max-plus algorithm to create a dynamic graph structure that represents the interactions among vehicles, emphasizing effective communication and coordination between CAVs. 

In \citep{he2022robust}, an innovative MADRL approach using constrained adversarial reinforcement learning to enhance the decision-making of autonomous vehicles at lane-based freeways with on-ramps merging is developed. By implementing this approach, the CAV’s agent gains the capability to make merging decisions with a high degree of robustness and safety, even in the presence of challenging environmental uncertainties and adversarial factors.\par
Moreover, a novel cooperative MADRL algorithm is introduced in \citep{chen2021graph}, that merges graphic convolution neural networks with deep Q-networks, resulting in an advanced graphic convolution Q network, serving as an information fusion module and decision processor. The trained model, in multi-lane road scenarios, enables CAVs to make efficient and safe lane-change decisions. This approach enhances both safety and mobility, ensuring CAVs can achieve their operational goals in partially observable and dynamic mixed-traffic environments. \par

While DRL and MADRL have seen extensive application in traditional lane-based traffic scenarios, there are limited applications of such approaches, specifically MADRL, in the context of lane-free driving. This work primarily considers a MADRL approach focusing on the dynamic and partially observable nature of such traffic environments that require novel definitions of environment, reward function, and network topology.

\section{Problem statement and formulation}\label{sec_3}
\subsection{Multi-agent deep reinforcement learning}\label{subsec_4_1}
Reinforcement Learning (RL) is a methodology in which an agent $p$ acquires an optimal policy, denoted as $\pi_{\theta_p}^*$, through a process of iterative trial and error \citep{sutton2018reinforcement}. This policy enables an agent to, by interacting with an environment, make a decision that maximizes its cumulative reward $R_{cum_p}=\sum\limits_{n=1}^T \gamma^nr_{p}(n)$ where $r_{p}(n)$ represents the reward function value associated with agent $p \in P$ at time step n, where $P$ is the total number of agents. Additionally, $\gamma$ and T denote a discount factor and the time horizon, respectively. Similar to human learning, agents in RL should construct and acquire knowledge directly from the input data. This is facilitated by DRL, which has introduced a range of value-based algorithms like Deep Q-networks (DQN) \citep{mnih2015human, quek2021deep}, and policy-based algorithms such as deep deterministic policy gradient (DDPG) \citep{silver2014deterministic, lillicrap2015continuous}. Deep neural networks are employed to approximate complex mappings from observations to actions, enabling the agent to generalize and handle high-dimensional input spaces. In this work, each vehicle represents an agent in the proposed algorithm\par
The traffic environment is non-stationary and partially observable from each agent's perspective, as each agent can only perceive local observations around itself. In such a dynamic environment, where agents aim to achieve individual objectives like maintaining their desired speed or performing overtaking maneuvers without considering other agents' goals, we observe a competitive interaction among these agents. Furthermore, it is imperative to emphasize that the actions of an agent could have an impact on the behavior of other agents, primarily due to the generation of nudging forces, i.e., an inter-vehicle force defined in lane-free traffic, imposed on the leading agents. Consequently, the resultant nudging forces influence the reward function values of the respective preceding agents, as will be explained later. This phenomenon holds significance across all consecutive agents.\par
Thus, there would be a competitive interaction between agents in such an environment. In light of the aforementioned fact, applying DQN and policy gradient algorithms in such multi-agent settings typically yields suboptimal results, as these methods do not effectively utilize information from other agents during training. To address this limitation, employing a MADDPG-based algorithm \citep{lowe2017multi} is recommended, which successfully overcomes this challenge by incorporating the states and actions of other agents during the training process.

The following sections explain the preliminary requirements for formulating CAV manipulation in a lane-free traffic environment as a MADRL problem.

\subsection{Lane-free traffic environment }\label{subsec_3_1}
As previously mentioned, the primary goal of this research is to develop a control strategy for CAVs operating in a traffic environment without designated lanes. In this unique setting, vehicles have the freedom to adjust their lateral positions without being restricted to fixed lanes. This allows more efficient exploitation of road lateral occupancy and enables vehicles for smooth and effective overtaking maneuvers.  Before delving into the details of the proposed strategy, it is important to understand the notion of a lane-free traffic environment, as introduced by Papageorgiou et al. (2021) via TrafficFluid concept \citep{papageorgiou2021lane}. TrafficFluid is founded on two core principles: "Lane-free traffic" and "Nudging." The first principle allows vehicles to maneuver freely across the entire width of the roadway, while the latter involves the application of a virtual pushing force by faster upstream vehicles on downstream vehicles to facilitate overtaking compared to traditional traffic, where vehicles are only influenced by the downstream traffic. While in a traditional lane-based traffic system, overtaking typically requires changing lanes, in a lane-free environment, nudging and repulsion, the force applied from a preceding vehicle to a following vehicle, and vice versa, can create the necessary space for overtaking.\par
\begin{figure}[h]
\centering
\includegraphics[width=0.5\textwidth]{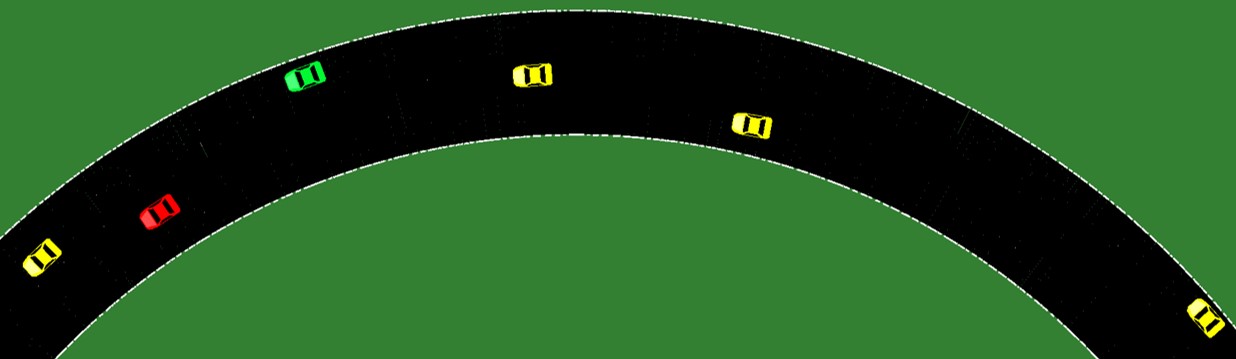}\\
\caption{Lane-free circular freeway used to train CAVs’ agent}
\label{fig_ringRoad}
\end{figure}

This study endeavors to develop a lane-free driving strategy through training multiple CAVs, hereinafter denoted as "agents". The training phase is carried out and evaluated on a circular roadway as illustrated in Fig.~\ref{fig_ringRoad} and is implemented in the same ring road at different traffic densities and a stretch of freeway with on- and off-ramps for further evaluations. Our proposed control methodology is intended to enable CAVs to perform overtaking maneuvers while avoiding collision; therefore, they could maintain their velocity next to their desired speed.

%\begin{graphicalabstract}
%\includegraphics{grabs}
%\end{graphicalabstract}

These agents will engage in dynamic interactions with one another, independently developing their best strategies while focusing on their unique objectives. This sort of interaction between agents results in the creation of a competitive multi-agent framework. Training agents in such a non-stationary environment equips us with the capacity to cultivate a robust control policy capable of adeptly accommodating the dynamic nature of the traffic environment. This adaptability is crucial for responding to shifts in neighboring vehicle behavior, which depend on factors such as the current environment and the agents' individual goals, including changes in their actual speeds and overtaking maneuvers.\par
It is imperative to establish certain foundational definitions to formulate the intended DRL algorithm. One such prerequisite involves defining an elliptical safety border surrounding each agent used to determine inter-vehicle forces, i.e., nudging and repulsion forces. Such definitions are clarified in the subsequent sections.\par

\subsection{Artificial Safety Border}\label{subsec_3_2}
\begin{figure}[h]
\centering
\includegraphics[width=0.5\textwidth]{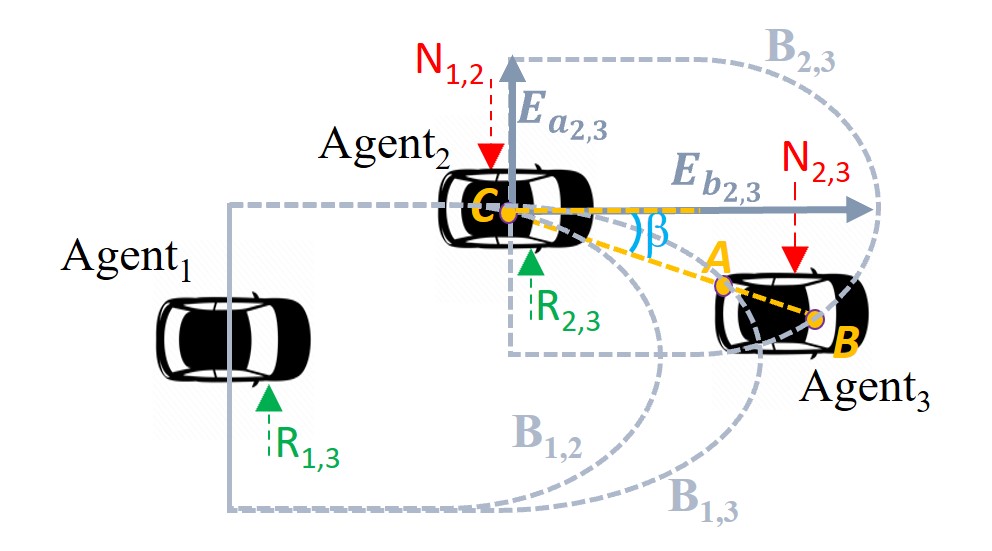}\\
\caption{Artificial semi-ellipses and their corresponding forces. N\textsubscript{1,2} and N\textsubscript{2,3} are lateral nudging forces applied to agent\textsubscript{2} and agent\textsubscript{3} from agent\textsubscript{1} and agent\textsubscript{2} respectively. R\textsubscript{1,3} and R\textsubscript{2,3} are lateral repulsive forces acting on agent\textsubscript{1} and agent\textsubscript{2} from agent\textsubscript{3}.}
\label{ellipsoide_borders}
\end{figure}

A prevalent method for assessing the likelihood of vehicle collisions in a lane-free traffic setting involves employing an elliptical border that acts as a safety zone around each vehicle \citep{troullinos2021collaborative, yanumula2021optimal, berahman2022driving}. This paper defines a semi-ellipse between each vehicle and any succeeding vehicle in its detection range, as shown in Fig. \ref{ellipsoide_borders}. The goal is to maintain a safe distance between vehicles by ensuring they remain outside the elliptical borders of others. Furthermore, these specified boundaries provide the structural foundation for introducing nudging and repulsive forces applied to respective vehicles, which are employed in CAVs' maneuvers, including overtaking and merging, as will be explained in the subsequent discussion.

In Fig.~\ref{ellipsoide_borders}, $B_{i,j}$ represents the artificial safety border of agent $i$ with respect to leading agent $j$. It is worth noting that the defined border is a dynamic area where the length of $E_{a_{i,j}}$ and $E_{b_{i,j}}$ (the semi-minor and semi-major axes of the ellipse) vary based on the relative lateral and longitudinal velocities of the corresponding vehicles. $E_{a_{i,j}}$ and $E_{b_{i,j}}$ should be sufficiently long, allowing the ego-vehicle to maintain an appropriate diagonal distance from its front surrounding vehicles, thereby enabling it to react effectively and avoid collisions when confronted with a variety of maneuvers done by the leading vehicles. However, maintaining excessive diagonal distances between vehicles can reduce the overall traffic throughput, which contradicts the fundamental principles of capacity increase introduced by the lane-free concept. Therefore, a balance must be struck between these two criteria when defining the aforesaid parameter.

In light of these considerations, $E_{b_{i,j}}(n)$, where $n$ is the discrete time index, is specified as follows:
\begin{equation}\label{eq_2}
    E_{b_{i,j}(n)} = d_{0_{lon}}+t_{ds}v_{lon_{i}}(n-1)+d_{v_{lon}}(n),
\end{equation}
where the variable $d_0$ depends on the length of the ego vehicle to ensure a minimum safe distance between vehicles, $v_{lon_{i}}(n-1)$ is the longitudinal velocity of the ego vehicle $i$ at time step $n-1$ and $t_{ds}$ represents the desired time headway \textemdash \ a crucial parameter defining a safe distance between the ego-vehicle $i$ and its leading vehicle $j$ driving at the same velocity. It is noteworthy to mention that the first and second terms in \eqref{eq_2} are similar to the required inter-vehicle distance in lane-based car-following models \citep{xiao2017realistic}. Additionally, the parameter $d_{v_{lon}}$ is introduced in this paper to account for speed differences between the ego vehicle and its leading vehicle. This parameter plays the role of an additional safety measure for situations where the speed of the ego vehicle is higher than the leading one, and overtaking is not possible. It represents the minimum distance the ego vehicle needs to slow down by applying its maximum deceleration to match the speed of its predecessor, that is driving at a lower speed, and is formulated as
\begin{equation}\label{eq_3}
%\begin{split}
d_{v_{lon}}(n) = 
\begin{cases}
.5dec_{max}t_{r_{i,j}}^2(n)+\Delta v_{i,j}(n)t_{r_{i,j}}(n) & \text{,}\; v_i(n) > v_{j}(n)\\\
0 & \text{,}\;  v_i(n) \leq v_{j}(n)
\end{cases}
%\end{split}
\end{equation}
where $dec_{max}$ represents the maximum deceleration of the ego vehicle, while $\Delta v_{i,j}(n)= v_{lon_i}(n) - v_{lon_j}(n)$ denotes the difference between the longitudinal speed of ego-vehicle $i$ and that of its leading vehicle $j$. $t_{r_{i,j}}(n)=\dfrac{\Delta v_{i,j}(n)}{dec_{max}}$ describes the time required for the ego vehicle to match its predecessor's speed while implementing its maximum deceleration. \par
Similarly, $E_{a_{i,j}}$ is also composed of two parameters. The first element, $d_{0_{lat}}$, is a width-dependant constant value, which is the minimum required lateral distance between the corresponding vehicles once they longitudinally overlap. The second term is a dynamic variable based on the relative lateral speed between the corresponding vehicles. Therefore, $E_{a_{i,j}}$  is defined as
\begin{equation}\label{eq_4}
	E_{a_{i,j}}(n) =d_{0_{lat}}+ d_{v_{lat}}(n),
\end{equation}
where $d_{v_{lat}}(n)$ is defined, in \eqref{eq_5}, in such a way as to widen the safety ellipse once two vehicles approach each other.
%\begin{figure*}[h]
\begin{equation}\label{eq_5}
\begin{split}
&d_{v_{lat}}(n) =\\
&\begin{cases}
-\sqrt{E_{b_{i,j}}(n)^2+d_{lon_{i,j}}(n)^2}\, \dfrac{\bar{d}_{v_{lat}}(n)}{E_{b_{i,j}}(n)} & \text{,} \quad \bar{d}_{v_{lat}}(n) <0 \: \& \: d_{lon_{i,j}}(n)<E_{b_{i,j}}(n)\\\
0 & \text{,} \quad  \text{otherwise}
\end{cases}
\end{split}
\end{equation}
%\end{figure*}

In \eqref{eq_5}, $d_{lon_{i,j}}(n)$ represents the longitudinal space gap between vehicles $i$ and $j$. This equation indicates that if two vehicles are not approaching or they do not intrude into their safety ellipses, the value of $d_{v_{lat}}(n)$ is zero; otherwise, a positive value is added to \eqref{eq_4} widening the safety border. $\bar{d}_{v_{lat}}(n)=d_{lat_{i,j}}(n)-d_{lat_{i,j}}(n-1)$, with $d_{lat_{i,j}}$ being the lateral distance between two vehicles $i$ and $j$, represents the change of the lateral gap in two consecutive time steps. Notably, it has a negative value when the two vehicles approach each other and a positive value when moving apart. \par
It is worth noting that a vehicle may have multiple ellipses (as depicted in Fig.~\ref{ellipsoide_borders} for agent\textsubscript{1}), each with a different coverage area which is determined using equations \ref{eq_2} - \ref{eq_5}.

The nudging and repulsive forces resulting from the vehicle’s border invasion are also illustrated in Fig.~\ref{ellipsoide_borders}. Note that for the sake of simplicity, only the lateral forces are shown. Nudging is a force applied to a preceding vehicle from succeeding vehicles, and repulsion is a force inserted from the vehicles downstream to the upstream ones. The intrusion of a vehicle $j$ into the defined safety border of a vehicle $i$ is determined by the following equation:

\begin{equation}\label{eq_1}
    q_{i,j}(n) = \frac{(x_A(n)-x_C(n))^2}{E_{b_{i,j}}^2(n)}+\frac{(y_A(n)-y_C(n))^2}{E_{a_{i,j}}^2(n)}
\end{equation}
where $x_C$ and $y_C$ denote the agent's longitudinal and lateral center positions, respectively (point C). The variables $ x_A$ ($\displaystyle{ x_A \geq x_C}$) and $ y_C$  describe the longitudinal and lateral positions of the nearest point on the preceding vehicle $j$, as observed from the perspective of agent $i$ (point A).  A $q_{i,j}\leq{1}$ indicates that the front agent is located within its following agent’s elliptical border. Once this intrusion is identified, the corresponding inter-vehicle force needs to be calculated based on the methodology in the following section.

\subsection{Calculation of nudge and repulsion forces}\label{subsec_3_4}
As previously mentioned, this study employs the concepts of nudging and repulsive forces to facilitate overtaking and merge and diverge maneuvers within the proposed algorithm. Once an intrusion is detected, the mentioned inter-vehicle force should be applied to vehicles such a way that the invasion is cleared by moving point $A$ to point $B$ as illustrated in Fig. \ref{ellipsoide_borders}. To this end, the calculation of point $B$ coordinate ($\displaystyle{x_B, \: y_B}$) is essential.\par
With point C serving as a reference point in Fig.~\ref{ellipsoide_borders}, and considering angle $\beta$, we can express $y_B(n)=x_B (n)\tan \beta(n)$. AS ($\displaystyle{x_B, \: y_B}$) satisfy the ellipse equation $B_{i,j}$, we have
\begin{equation}\label{eq_7}
\frac{x_B^2(n)}{E_{b_{i,j}}^2(n)}+ \frac{y_B^2(n)}{E_{a_{i,j}}^2(n)}=1.
\end{equation}
Then, by substituting $y_B$ with $x_B\tan \beta$ in \ref{eq_7},  we can obtain the value of $x_B$ using  
\begin{equation}\label{eq_8}
x_B(n)=\sqrt{\frac{E_{b_{i,j}}^2(n)E_{a_{i,j}}^2(n)}{E_{b_{i,j}}^2(n)\tan^2\beta(n) + E_{a_{i,j}}^2(n)}}.
\end{equation}
Finally, by substituting the value of $x_B$ into \ref{eq_7}, the value of $y_B$ is determined.

In the next step, the intrusion percent, $IntPer_{i,j}$, is defined as the intrusion percentage of $B_{i,j}$ by its corresponding leading vehicle $j$ which is computed as follows:
\begin{equation}\label{eq_6}
IntPer_{i,j} =\frac{\lVert AB_{i,j} \rVert}{\lVert CB_{i,j} \rVert}
\end{equation}
where  $\lVert AB_{i,j} \rVert$ and $\lVert CB_{i,j} \rVert$ represent the Euclidean distances between point A and point B, and between point C and point B, respectively, for the two agents $i$ and $j$ (as depicted in Fig.~\ref{ellipsoide_borders}).\par

We propose that the repulsion force $F_{rep_{i,j}}$, applied to the vehicle $i$ from the leading vehicle $j$, be equal to $IntPer_{i,j}$. Additionally, a portion of this value is used for the calculation of the nudging force $F_{nud_{i,j}}$ exerted on the corresponding preceding vehicle $j$, which is determined by
\begin{equation}\label{eq_9}
F_{nud_{i,j}}(n)=\alpha IntPer_{i,j}(n) + F_{s,j}(n), 
\end{equation}
where $\alpha$ is a constant design coefficient. There are some cases where, due to the interaction of repulsion and nudging forces, the leading vehicle $j$ remains on the border, and therefore, no intrusion is detected. In such cases, as no new force is being generated, the following vehicle $i$ is unable to overtake and may drive at a lower speed compared to its desired speed, becoming locked to follow the leading vehicle. The second term in Equation \ref{eq_9} is introduced to allow the following vehicle to speed up and overtake, providing a stronger nudging force. $F_{s,j}$ is determined as
\begin{equation}\label{eq_10}
F_{s,j}(n) = \frac{S_{d_i}(n)}{1+\alpha IntPer_{i,j}(n)},
\end{equation}
where $S_{d_i}(n)=\max(0,s_{d_i})$, with 
\begin{equation}\label{eq_11}
s_{d_i} = \frac{v_{d_i}(n)-v_{lon_i}(n)}{v_{d_i}(n)}.
\end{equation}

Therefore, if the actual speed $v_{lon_i}$ is less than the desired speed $v_{d_i}$, an additional nudging force is applied to the leading vehicle, making more space for the following vehicle to overtake. Note that based on the denominator of \ref{eq_10}, the value of this additional term has the highest value when there is no intrusion and is weakened by more intrusion of the safety border.

It should be clarified that, as illustrated in Fig.~\ref{ellipsoide_borders}, any vehicle $i$ might receive several nudging and repulsive forces from surrounding vehicles. However, we only assume the maximum values of the forces as $F_{nud_i}$ and $F_{rep_i}$ to account for the worst-case conditions. Section \ref{sub_subsec_3_3_3} explains how these forces are used in the agent's reward function. The reward function is defined in such a way that the impact of the nudging force is a lateral movement to the right, and the effect of a repulsive force is slowing down and moving to the left. In contrast to the original lane-free concept that allows overtaking from both sides, this innovative approach eliminates unnecessarily lateral movements by leading faster vehicles to the left and slow vehicles to the right side of the road. Consequently, nudging and repulsive forces manipulate vehicles' trajectories, minimize intrusion, or ideally prevent collisions, allowing smooth overtaking and a laminar traffic flow.\par
The lateral response of each agent to the imposed repulsive and nudging force provides an opportunity to conduct merging and diverging maneuvers in a lane-free road with on- and off-ramps where each vehicle has its own designated driving route. Following such routes is one of the imperative objectives of this work. Extra lanes \textemdash acceleration and deceleration lanes on the right side of the road\textemdash are used to facilitate these mentioned maneuvers. When an agent engages in a merging maneuver from an on-ramp and navigates along the acceleration lane, a consistent virtual lateral repulsive force is implemented to steer it toward the mainstream traffic. The agent responds by laterally moving to the left, reducing the force to increase the received reward. Once the agent successfully merges onto the main road, the virtual repulsive force is removed. \par
During a diverging maneuver at an off-ramp, a constant virtual nudging force is applied to guide the intended vehicle toward the right side of the road, facilitating entry into the deceleration lane. However, due to the ellipsoid borders of the intended agents, they may encounter intrusion by their predecessors, resulting in repulsive forces. As discussed earlier, the algorithm's response to these forces may induce leftward movement, contrary to the desired rightward direction needed for exiting at off-ramps.
To counteract this issue, any action causing leftward movement during diverging at off-ramps is suppressed. Consequently, the presence of repulsive forces from such vehicles only results in speed reduction in the intended agent to prevent collisions. Once the agent enters the deceleration lane, the virtual nudging force is discontinued, allowing the agent to follow the deceleration lane and exit the road.
It's important to note that departing agents receive the virtual repulsion force upstream of the deceleration lane, enabling them to transition to the right side of the road before entering the deceleration lane. This early diverging maneuver ensures that departing agents have ample time and longitudinal space to execute the diverging maneuver, particularly in congested traffic situations.

\subsection{Problem formulation in Markov Decision Process framework}\label{subsec_3_3}
After the necessary developments for the movement of vehicles in the lane-free environment, we now delve into the discussion of the definitions needed in the proposed DRL-based algorithm and its training approach. Developing such an algorithm needs to be represented as a Markov decision process (MDP) \citep{sutton2018reinforcement}. To this end, it is necessary to define the state space, action space, and reward function for each agent, as will be discussed in the following sections.\par

\subsubsection{State Space definition}\label{sub_subsec_3_3_1}
In order to control the maneuvers of CAVs based on the decisions made by their respective agents, it is crucial to have a clear understanding of the current state of CAVs within their traffic environment. Considering the primary objective of each agent, collision avoidance and driving at the desired speed, the state space encompasses parameters such as $s_{d}$, defined in \eqref{eq_11}) and the actual longitudinal speed of the vehicle ($v_{lon}$).
Additionally, to account for passenger comfort during vehicle maneuvers, longitudinal acceleration ($acc_{lon}$) and lateral speed ($v_{lat}$) of the vehicle are also considered in the state space.\par
\begin{figure}[H]
\centering
\includegraphics[width=0.4\textwidth]{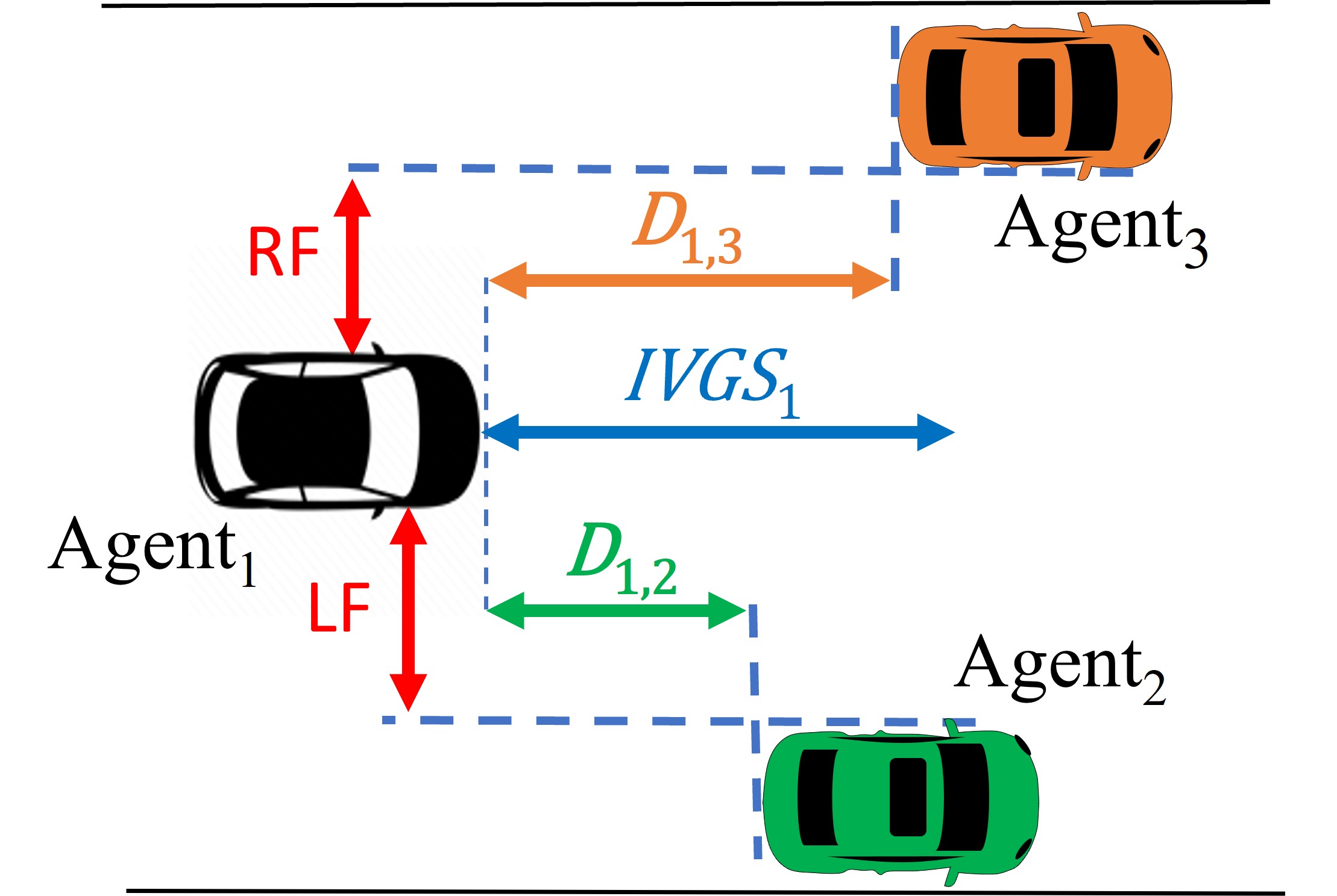}\\
\caption{Definition of left and right freedom, $fr_l$ and $fr_r$, for CAVs in lane-free traffic freeway.}
\label{IVGS}
\end{figure}

Moreover, we introduce left and right lateral freedom ($fr_l$, $fr_r$) as additional state space elements. $fr_l$ and $fr_r$ provide agents with information about their current lateral conditions in their vicinity, allowing them to make more precise actions. 
To this end, we first use the concept of inter-vehicle gap set point defined as 
\begin{equation}\label{eq_12}
IVGS_i (n)= d_{0_{lon}}+t_{ds}v_{lon_{i}}(n-1),
\end{equation}
where $IVGS_i$ represents the desired longitudinal space gap between vehicle $i$ and its front vehicles, with $d_{0_{lon}}$, $t_{ds}$, and $v_i$ having the same definitions as used in \eqref{eq_2}. The value of $IVGS_i$ is employed in determining $fr_l$ and $fr_r$ as depicted in Fig.~\ref{IVGS}.

The lateral gap is computed for leading vehicles, or agents, whose longitudinal space gap concerning vehicle $i$ is less than $IVGS_i$. The minimum lateral gap to the vehicles on the left is denoted as $fr_l$, and the minimum lateral gap to the right vehicles is $fr_r$. 

Obviously, if there is no preceding agent, the lateral gaps are measured with respect to the road boundaries. Notably, the definition of lateral freedom indicators, $fr_l$ and $fr_r$, used later in the reward function, plays the role of an additional safety measure preventing collision or leaving the road boundaries.\par

Last but not least, the repulsive and nudge forces applied on the vehicle, defined in Section \ref{subsec_3_4}, are also included in the state space of each agent. 

To recap briefly, the state space ${S_p}$ for each agent $p$ is comprised of: $s^p_{d}$, $v^p_{lon}$, $acc^p_{lon}$, $v^p_{lat}$, $fr^p_{l}$, $fr^p_{r}$, $F^p_{rep}$, and $F^p_{nud}$. The proposed DRL algorithm uses the state space $\bm{S}$ for all agents to determine the required actions, maximizing the reward value discussed in the subsequent section.
\subsubsection{Action Space}\label{sub_subsec_3_3_2}
Given the principles of the driving strategy in LFT, two actions are considered for each agent, i.e., longitudinal acceleration and lateral speed. The longitudinal acceleration for each agent, $acc_{lon}$ takes a continuous value bounded by a predetermined interval of [$-acc_{max}$, $acc_{max}$]. Note that based on this definition, a negative acceleration indicates braking. In addition, each agent, by changing its lateral speed, $v_{lat}$, can manipulate its lateral position. It is important to note that a positive value of $v_{lat}$ corresponds to a leftward lateral movement on the road, while a negative value represents a rightward lateral adjustment. Additionally, the range of $v_{lat}$ is also limited to [$-v_{lat_{max}}$, $v_{lat_{max}}$].\par
Building upon this definition, the action space $A$ for each agent comprises $acc_{lat}$ and $v_{lat}$. The value of these parameters is determined dynamically at each time step by the proposed DRL-based algorithm, which is trained to maximize its reward function.Therefore, actions $(acc^p_{lat}, v^p_{lat})$ of each agent $p$ is defined by $a_p\in {A}$.

\subsubsection{Reward function}\label{sub_subsec_3_3_3}
The reward function is the core of the proposed methodology, similar to any other learning-based algorithm, as it defines the agents' behavior. It also influences the convergence and speed of convergence in a DRL algorithm. Based on the proposed control strategy, the training of the DRL algorithm in this work is carried out based on a multi-objective problem embedded within the reward function. 

The first objective is facilitating overtaking maneuvers and preventing collision via the applied repulsive force $F_{rep}$. Thus, the first term in the reward function is defined as follows:
\begin{equation}\label{eq_13}
r_{rep}(n)=-F_{rep}(n).
\end{equation}
Since the training algorithm always tries to maximize the reward value, this term in the reward function leads to a reduction of repulsive force, keeping distance from the front vehicles by either decelerating or moving to the right.
In addition to collision avoidance behavior due to repulsion, we should also allow enough moving forward power, leading to increasing the traffic throughput and driving close to desired speeds. Hence, the second element of the reward function is introduced as:
\begin{equation}\label{eq_14}
r_{nud}(n)=-w_{nud}(n)\big(s_d(n)+F_{nud}(n)\big),
\end{equation}
where  $w_{nud}$, defined in \eqref{eq_15}, is a weighting factor for this reward term, making a trade-off between the effect of nudging and repulsion.
\begin{equation}\label{eq_15}
w_{nud}(n)=\max\bigg(0,\: 1-\frac{F_{rep(n)}}{F_{rep_{t}}}\bigg).
\end{equation}
Here, $F_{rep_{t}}$ is a constant design parameter that defines a threshold for the repulsion force. Given \eqref{eq_15}, any repulsion force applied to the vehicles larger than this threshold sets the nudging term in the reward function, \eqref{eq_14}, to zero, i.e., giving higher priority to collision avoidance in the agent's action policy over other objectives like maintaining the desired speed or reducing nudging force. In other words, once there is enough nudging force applied to a vehicle from the vehicle behind, the agent tends to either increase its speed or move to the right. But if, at the same time, there is an  applied repulsion from the vehicles in front, the moving forward temptation is reduced by making $s_d$ and $F_{nud}$ less effective in the reward function calculation. It is crucially important in dense traffic where vehicles should demonstrate less aggressive actions as there might not be enough space downstream.  Consequently, $F_{rep}$ takes precedence in the determination of the reward function.\par
Furthermore, to incorporate passenger comfort in the design of the DRL algorithm, impacts of longitudinal jerk and lateral acceleration are introduced in the reward function. Note that, in the longitudinal direction, the most comfortable situation is driving at a constant speed, i.e., zero acceleration, or with steady acceleration. Therefore, any change in the acceleration, i.e., jerk, is considered the metric for longitudinal comfort. While in the lateral direction, the ideal situation is maintaining the current lateral location, i.e., lateral zero speed, or moving to the side with constant lateral speed. Therefore, the change in the lateral speed, i.e., lateral acceleration, represents the comfort in this direction.

To this end, the longitudinal jerk reward defined in \citep{zhu2020safe} is utilized to incentivize smooth longitudinal movement of each agent, which is formulated as follows:
\begin{equation}\label{eq_16}
r_{J_{lon}}(n)=-w_{jer} \Delta t\frac{| J_{lon}(n)| }{\Delta acc_{max}},
\end{equation} 
where $w_{jer}$ is a constant weighting coefficient determining the intensity of the jerk effect on the overall reward computation, $\Delta t$ is the time step, and  $\Delta acc_{max}$ is the maximum acceleration change, i.e., the difference between the maximum acceleration and maximum deceleration, that an agent could take (here equals to $2\cdot acc_{max}$). Moreover, $J_{lon}$, the longitudinal jerk, is determined as
\begin{equation}\label{eq_17}
J_{lon}(n)=\frac{acc_{lon}(n)-acc_{lon}(n-1)}{\Delta t}.
\end{equation}

The lateral reward ($r_{acc_{lat}}$) is introduced as
\begin{equation}\label{eq_18}
r_{acc_{lat}}(n)=-w_{acc} \Delta t\frac{|acc_{lat}(n)| }{\Delta v_{lat_{max}}}.
\end{equation}
In \eqref{eq_18}, $w_{acc}$ is a weighting factor for lateral acceleration reward, $\Delta v_{lat_{max}}$ denotes the maximum possible lateral speed variation between two consecutive time steps (here equals to $2\cdot v_{lat_{max}}$), and $acc_{lat}$ is the lateral acceleration defined as follows:
\begin{equation}\label{eq_19}
acc_{lat}(n)=\frac{v_{lat}(n) - v_{lat}(n-1)}{\Delta t}.
\end{equation}

Finally, yet importantly, we penalize incorrect lateral actions taken by the agents to speed up the convergence of the developed algorithm toward the optimal action policy. For instance, any decision by an agent resulting in lateral movement beyond the defined left and right freedoms ($fr_l$, $fr_r$) is deemed incorrect. Furthermore, we have structured our strategy so that a nudge force propels the slow vehicle to the right, and repulsion guides it to the left. To implement this, we evaluate the maximum of the nudge and repulsion forces applied to each vehicle and compare it with the lateral action recommended by the agent. If, for example, the nudge is greater than the repulsion, a lateral action to the left incurs a penalty. If no wrong action is identified, this penalty term $r_{pen_{lat}}$ remains at zero.\par
Given the different terms of the reward function defined in \ref{eq_13} - \ref{eq_19}, we derive the ultimate value of the reward function $r_p(n)$ at time step $n$ for each agent $p \in P$ where $P$ is the total number of agents as follows:
\begin{equation}\label{eq_20}
\begin{split}
r_p(n)= r^p_{rep}(n) \; &+ \; w_{nud}r^p_{nud}(n) \; + \; w_{jer}r^p_{J_{lon}}(n) \\ &+ \; w_{acc}r^p_{acc_{lat}}(n) \; + \; r^p_{pen_{lat}}(n)
\end{split}
\end{equation}
The value of reward function $r_p(n)$ is determined by evaluating the action $a_{p}(n)\in {A}$ taken by the agent $p$ with the environment observation $s_{p}(n)\in {S}$.
\section{Proposed multi-agent deep reinforcement learning algorithm}\label{sec_4}
As mentioned earlier, we observe a competitive interaction among the agents in the traffic environment, as each agent merely seeks to maximize its own reward function value. Furthermore, regarding the reward function defined in \eqref{eq_20}, it is imperative to emphasize that an agent's actions could impact the rewards of other agents due to the effect of nudging and repulsive forces. Therefore, we developed and employed a multi-agent reinforcement learning approach in this work, which is elaborated in this section.

The framework and workflow of the proposed MADDPG algorithm are illustrated in Fig. 4. Each agent is treated as a DDPG agent, with the distinction that states and actions are jointly shared among agents during training. Each agent comprises two networks: an actor network \scalebox{1.15}{$\mu_p $} and a critic network \scalebox{1.15}{$Q_p $}.\par
The computation of action $a$ at each time is done by its agent actor network relying solely on its local current state $s$, while in the training phase, the critic network employs current states and actions of all agents ($\mathbf{s}, \mathbf{a}$) to evaluate the local action $a$, where $\mathbf{s}=(s_1, s_2,\dotsc,s_P)$, $\mathbf{a}=(a_1, a_2,\dotsc,a_P)$.
Implementing this approach maintains a stationary environment throughout training because each agent's critic network is aware of the actions and state of all agents. During the execution phase, only the actor networks are active, establishing MADDPG as a framework that centralizes training and decentralizes execution. The main networks \scalebox{1.15}{$\mu_p $} and \scalebox{1.15}{$Q_p $} are initialized with random weights, aimed at promoting training stability. Additionally, for further stability in the training process, target actor networks denoted as \scalebox{1.15}{$\mu_p' $}, and target critic networks, denoted as \scalebox{1.15}{$Q_p' $}, which are identical to the main networks \scalebox{1.15}{$\mu_p $} and \scalebox{1.15}{$Q_p $} are created. The weights of the target networks are initialized and updated to match those of the main networks, such that $\theta_p^{\mu_p}\to \theta_p^{\mu_p'}$ and $\theta_p^{Q_p}\to \theta_p^{Q_p'}$.\par
\begin{figure}
\centering
\includegraphics[width=1\textwidth]{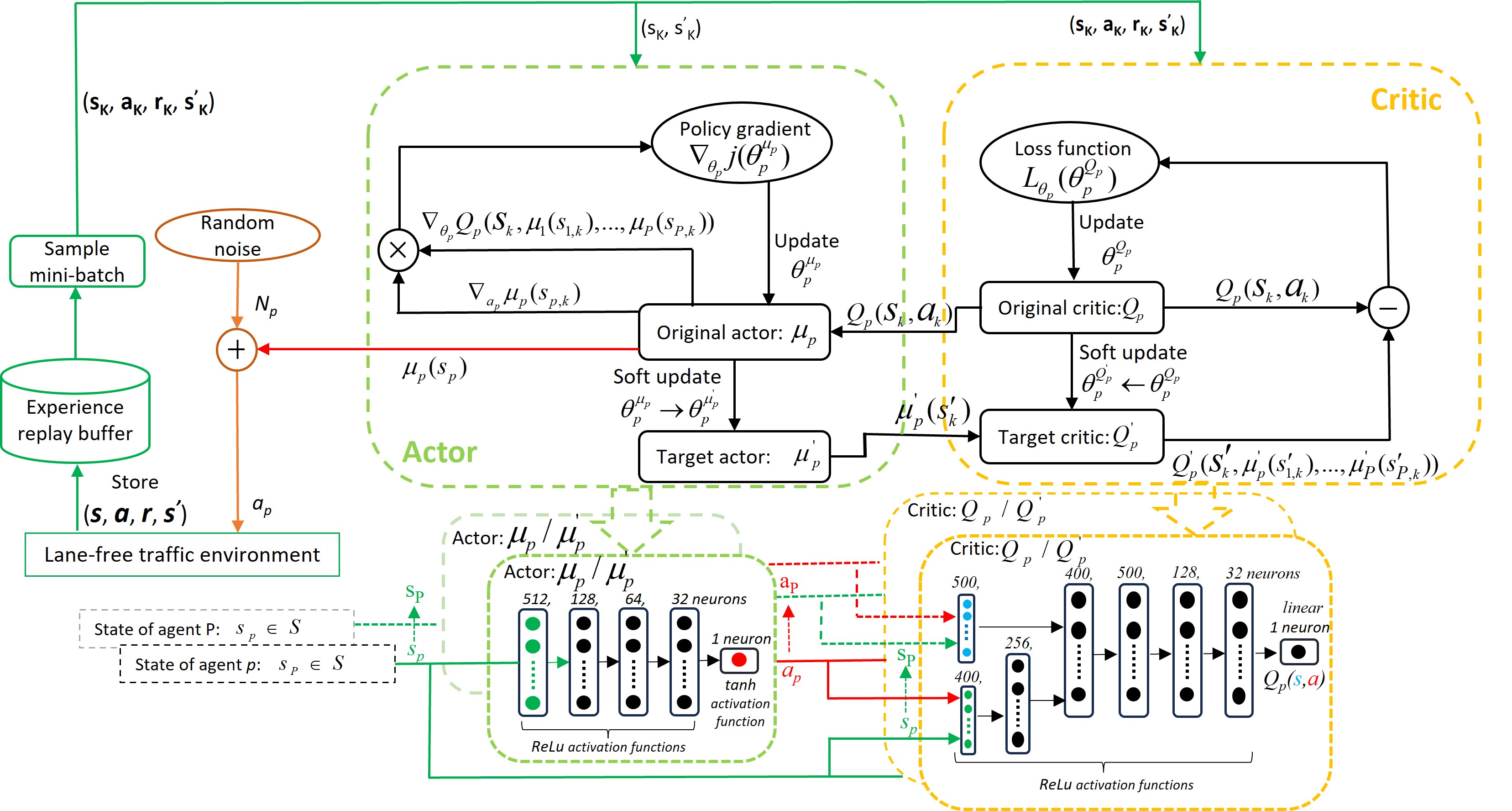}\\
\caption{The structure and operational sequence of the proposed MADDPG algorithm. Each agent, denoted as $p$, where $p=1,2,\dotsc ,P$, comprises two primary components: an original actor network \scalebox{1.15}{$\mu_p $}, with a corresponding target actor network \scalebox{1.15}{$\mu_p' $}, as well as an original critic network \scalebox{1.15}{$Q_p $}, accompanied by a target critic network \scalebox{1.15}{$Q_p' $}}
\label{MADDPG}
\end{figure}
Moreover, a replay buffer is utilized to store current states observation, actions and rewards belonging to all throughput agents along with the agents' next states observation, denoted as ($\mathbf{s}, \mathbf{a}, \mathbf{r}, \mathbf{s}^{'}$)  known as experience, where $\mathbf{r}=(r_1, r_2,\dotsc,r_P)$, $\mathbf{s}^{'}=(s^{'}_1, s^{'}_2,\dotsc,s^{'}_P)$.. The utilization of a replay buffer enhances training stability within the MADDPG framework, facilitating the agents in learning their optimal policies by selecting random mini-batches from the entirety of collected experiences.\par
In each training episode, agents are initialized with random states. A stochastic action exploration process, incorporating random noise generated by the Ornstein–Uhlenbeck process \citep{uhlenbeck1930theory} (denoted as $N_{p}$, ), is employed to facilitate action exploration in the early episodes of the training process.
Regarding the current state of agent $p$ and its generated noise at that time $N_{p}$, action is made by the agent through
\begin{equation}\label{eq_21}
a_{p}=\mu_p(s_{p})+ N_{p}
\end{equation}
here $\mu_p(s_{p})$ represents the action output belonging to \scalebox{1.15}{$\mu_p $}  network. Executing the resultant action for each agent transitions them to the next state, denoted as $s^{'}_{p}$. The agent's reward is subsequently calculated by evaluating this new state using \eqref{eq_20}. This sequence of actions is carried out for all participating agents. Following this, the collective experience ($\mathbf{s}, \mathbf{r}, \mathbf{a}, \mathbf{s}'$) is stored in the replay buffer and the current state get updated $\mathbf{s}\to \mathbf{s}^{'}$.
After the $\eta$ time step the neuron weights of the main actor and the main critic network for each agent $p$ are updated by means of a mini-batch comprised of $K$ samples extracted from the replay buffer in the training process.  Each agent $p$ updates the weights of its main critic network (i.e., $\theta_p^{Q_p}$) by means of Bellman’s principle of optimality, employing the gradient descent optimization algorithm, to minimize the mean-squared loss that is formulated as:
\begin{equation}\label{eq_22}
L(\theta_p^{Q_p})=\frac{1}{K}\sum\limits_{k=1}^K(y_{p,k}-Q_p(\mathbf{s}_k,\mathbf{a}_k))^2
\end{equation}
Regarding each member of the mini-batch samples as ($\mathbf{s}_k, \mathbf{a}_k, \mathbf{r}_k, \mathbf{s}_k'$), in \eqref{eq_22}, $Q_p(\mathbf{s},\mathbf{a}_k)$ represents the predicted output of the main critic network, and \scalebox{1.15}{$y_{p,k}$} denotes its target value, which is determined by
\begin{equation}\label{eq_23}
y_{p,k}=r_{p,k}+ \gamma Q_p'(\mathbf{s}_k', a_{1,k}',\dotsc,a_{P,k}')
\end{equation}
where $a_{p,k}'=\mu_{p}'(s_{p,k}')$  is the predicted action made by target actor network \scalebox{1.15}{$\mu_p'$} belonging to agent $p$ with respect to $s_{p,k}'$, representing the next state of agent $p$ in the $k$th sample of the selected mini batch; furthermore, \scalebox{1.15}{$Q_p'$} is the target network of agent $p$.\par
The weight parameters of the actor networks (i.e., $\theta_p^{\mu_p}$) are updated in a direction that maximizes their corresponding \scalebox{1.15}{$Q_p$} value by performing the gradient ascent optimization algorithm as follows:
\begin{equation}\label{eq_24}
\nabla_{\theta_p^{\mu_p}}k(\theta_p^{\mu_p})=\nabla_{\theta_p^{\mu_p}}\mu_p(s_{p,k})\: \nabla_{a_{p,k}}Q_p(\mathbf{s}_k,\mathbf{a}_k)
\end{equation}
where $\mathbf{a}_k=(\mu_1(s_{1,k}),\dotsc,\mu_1(s_{P,k}))$. Moreover, the weight parameters of target actor and critic networks (i.e., $\theta_p^{\mu_p'}$, $\theta_p^{Q_p'}$) are updated as follows.
\begin{equation}\label{eq_25}
\theta_p^{\mu_p'}=\tau\theta_p^{\mu_p}+(1-\tau)\theta_p^{\mu_p'}
\end{equation}
\begin{equation}\label{eq_26}
\theta_p^{Q_p'}=\tau\theta_p^{Q_p}+(1-\tau)\theta_p^{Q_p'}
\end{equation}
$\tau$ in (\ref{eq_25}, \ref{eq_26}) denotes the update rate defining the effect of the main actor and critic networks on their target counterparts.
The convergence of the proposed MADDRL algorithm leads to obtaining an optimal actor network for each involved agent regarded as \scalebox{1.15}{$\mu_p^*$}. Thereafter, each agent $p$ by merely observing its local state $s_{p,n}$ at current time step n obtains its optimal actions for the next execution $a_{p,n+1}=\mu_p^*(s_{p,n})$.
\section{Simulation set up and performance analysis}\label{sec_5}
The proposed methodology is evaluated across diverse scenarios, and the results are presented in this section. We start by describing the setup for the DRL algorithms and simulation environment for the training phase. Subsequently, we present the results of testing the proposed approach in two different networks and engage in discussion.
\subsection{Setup for the DRL Algorithm}\label{subsec_5_1}
In this work, each agent's actor and critic networks are constructed employing the hyper-parameters listed in \autoref{table_1}. The MADDPG algorithm is implemented in Python, utilizing the PyTorch framework \citep{paszke2019pytorch}. Weight coefficients of the networks were updated at each learning step using the Adam optimization method \citep{kingma2015adam}. Action selection during training followed the \scalebox{1.1}{$\epsilon$}-greedy policy, allowing a balanced approach to exploration and exploitation. \scalebox{1.1}{$\epsilon$} linearly decreased from 1 (indicating 100\% exploration) to 0.1 (indicating 10\% exploration) over the initial 200 episodes, and remained fixed at 0.1 thereafter. Furthermore, ReLU activation functions \citep{arora2018understanding} are employed for all actor and critic hidden layers. The actor networks output layer used the hyperbolic tangent ($\tanh$) activation function to yield a vector of continuous values within the range [-1, 1], while the critic networks utilized a linear activation unit in its output layer. Therefore, the acceleration/deceleration of each agent is obtained by mapping the output value of the actor network to the range of the predefined maximum acceleration/deceleration values.\par
\begin{table}[ht!]
\fontsize{7pt}{7pt}\selectfont
\centering
\caption{Hyper-parameter settings for actor and critic networks.}
\def\arraystretch{1.2}
\begin{tabular}{m{2cm}|m{1.2cm}|m{2.cm}|m{1.3cm}}
\hline
%https://www.baeldung.com/cs/latex-tables-wrap-text
%\RaggedRight{}\newline
\Centering{\textbf{Parameter}} & \Centering{\textbf{value}} & \Centering{\textbf{Parameter}} & \Centering{\textbf{value}} \\
 \hline
 \Centering{Discount factor ($\gamma$)}&  \Centering{0.95} & \Centering{Update rate ($\tau$)} & \Centering{0.01}\\
 \hline 
  \Centering{Batch size} & \Centering{128} & \Centering{No. of actor hidden layers} & \Centering{3}\\
  \hline
  \Centering{Actor \& Critic optimizer} & \Centering{Adam} & \Centering{No. of critic hidden layers} & \Centering{5}\\
   \hline
  \Centering{Actor \& Critic \newline{learning rate}} & \Centering{$1\times 10^{-4}$\newline ($1\times 10^{-3}$)} & \Centering{No. of nodes in each actor layer} & \Centering{512, 128, 64, 32,1}\\
 \hline
   \Centering{Experience replay buffer size} & \Centering{$1\times 10^6$} & \Centering{No. of nodes in each critic layer} & \Centering{500, 400, 256, 400, 500, 128, 32, 1}\\
 \hline
\end{tabular}
\label{table_1}
\end{table}
To train the proposed MADDPG algorithm and assess its performance in the lane-free driving concept, various lane-free scenarios were generated using the Simulation of Urban Mobility (SUMO) \citep{lopez2018microscopic}. As SUMO does not inherently support the implementation of lane-free traffic scenarios, this study leverages the sublane feature to address this gap.
By employing this feature, each lane is subdivided into multiple sublanes, allowing lateral movement of vehicles within the lane. This involves modifying the parameter associated with the lateral resolution ($L_r$) of a lane in SUMO to a low value, thereby creating multiple sublanes within a single lane. This configuration facilitates smooth lateral vehicle movements, a crucial element in a lane-free environment.

In line with the lane-free requirements of these scenarios, specific constraints within SUMO, designed for a lane-based traffic environment, have been disabled. For instance, there is no need to adhere to speed behavior restrictions prescribed by SUMO for vehicles sharing the same lane. In the lane-free context, a wide lane is assumed, comprising numerous sublanes, permitting vehicles to maintain close longitudinal distances as long as their lateral positions within the related sublanes are appropriately spaced.
Through the application of the aforementioned adjustments within SUMO, vehicles can be situated at nearly any arbitrary lateral position within a road boundary consisting of a single lane, as visually depicted in Fig.~\ref{fig_ringRoad}.\par
Furthermore, to enact the envisaged driving methodology within a traffic scenario, it is imperative to establish an interface for seamless communication with SUMO. In this regard, SUMO provides the necessary functionality through the Traffic Control Interface (TraCI) \citep{wegener2008traci}, an Application Programming Interface (API) that facilitates access to the attributes of simulated entities and enables the manipulation of their behavior. This architecture is used in the agent training phase and testing the trained agent in different scenarios, as will be explained in the following. \par

\subsection{Evaluation of the training phase}\label{subsec_5_2}
This section provides an analysis of convergence and performance for the proposed algorithm. The training of the developed algorithm is conducted by considering a multi-agent environment comprising six agents, navigating a lane-free ring road as depicted in Fig.~\ref{fig_ringRoad}. In every training episode, agents are initiated at predetermined locations on this road, each assigned distinctive initial and target speeds selected randomly from a uniform distribution ranging from 25 m/s to 35 m/s. Each learning episode terminates either when a collision between agents occurs or 1000 simulation steps have elapsed; the related simulation parameters are given in \autoref{table_2}.\par

\begin{figure}[ht!]
    %\centering
    \begin{subfigure}[t]{0.48\textwidth}
        \includegraphics[width=\textwidth]{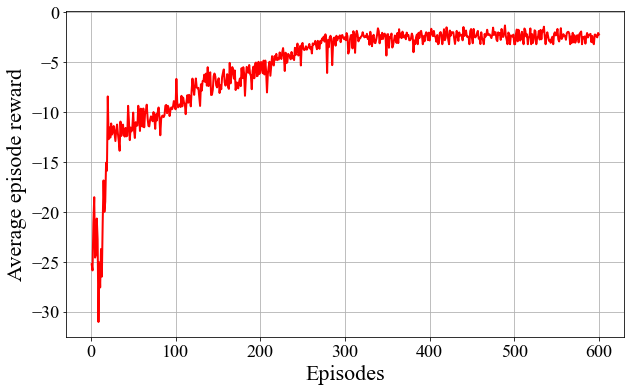}
        \caption{Average episode reward}
        \label{R_curve}
    \end{subfigure}
    \begin{subfigure}[t]{0.48\textwidth}
        \includegraphics[width=\textwidth]{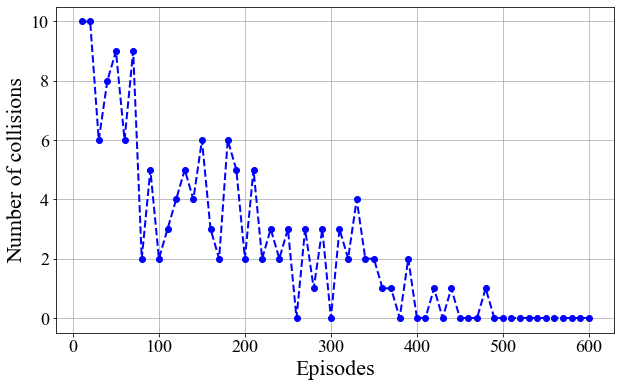}
        \caption{Each point represents the number of collision occurrences at each 10-episode interval}
        \label{col_Num}
    \end{subfigure}
 \caption{Average episode reward and collision occurrences  during the MADDPG training process}
\label{MADDPG_Training}
\end{figure}

The agents undergo training through a total of 600 episodes, and the progression of average episode reward $\bar{R}$, formulated in \eqref{eq_27}, is visualized in Fig.~\ref{R_curve}.
\begin{equation}\label{eq_27}
\bar{R}=\frac{1}{E_T}\sum\limits_{n=1}^{E_T}\sum\limits_{p=1}^Pr_{p}(n),
\end{equation}
where $E_T$ and $r_{p}$ represent the total elapsed simulation step at each episode and the value of reward function computed through \eqref{eq_20} associated with agent $p$ at time step n, respectively.\par

\begin{table}[h]
\fontsize{7pt}{7pt}\selectfont
\centering
\caption{Simulation and reward parameters.}
\def\arraystretch{1.2}
\begin{tabular}{m{2cm}|m{1.2cm}|m{2.cm}|m{1.3cm}}
\hline
%https://www.baeldung.com/cs/latex-tables-wrap-text
%\RaggedRight{}\newline
\Centering{\textbf{Parameter}} & \Centering{\textbf{value}} & \Centering{\textbf{Parameter}} & \Centering{\textbf{value}} \\
 \hline
 \Centering{Simulation time-interval
(\scalebox{1.2}{$\Delta t$)}}&  \Centering{0.25s} & \Centering{Desired and initial speed range} & \Centering{$[25,35]$ m/s}\\
 \hline 
  \Centering{Ring road length (width) } & \Centering{400 (10.2) m} & \Centering{Lateral resolution (sublane width)} & \Centering{0.2 m}\\
  \hline
  \Centering{Maximum acceleration (deceleration)} & \Centering{4 (-4) m/s\textsuperscript{2}} & \Centering{Maximum lateral speed} & \Centering{1.5 m/s}\\
   \hline
  \Centering{Agent’s length (width)} & \Centering{3.2 (1.8) m} & \Centering{Maximum simulation step in training phase} & \Centering{1000}\\
 \hline
   \Centering{No. of agents in training phase (\textit{P})} & \Centering{6} & \Centering{Desired time headway(\scalebox{1.2}{$t_{ds}$})} & \Centering{0.5s}\\
 \hline
    \Centering{lateral reward punishment (\scalebox{1.2}{$r_{pen_{lat}}$})} & \Centering{-5} & \Centering{lateral jerk coef. (\scalebox{1.2}{$\xi$})} & \Centering{0.4}\\
 \hline
     \Centering{$F_{rep_t}$ in \eqref{eq_15}} & \Centering{0.3} & \Centering{longitudinal jerk coef. (\scalebox{1.2}{$\delta$})} & \Centering{0.4}\\
 \hline
\end{tabular}
\label{table_2}
\end{table}
The convergence of the developed MADDPG algorithm is depicted in Fig.~\ref{MADDPG_Training}a. It indicates that the average reward converges relatively fast and stabilizes around -3 after 500 episodes. Notably, no collisions occurred beyond this point, demonstrating that the trained algorithm successfully provides a collision-free circumstance.
Since one of the primary objectives of the proposed MADDPG algorithm is to ensure collision-free conditions, the number of collisions within 10 episode intervals during the training phase is counted and illustrated in Fig.~\ref{col_Num}, a critical metric for validating the efficacy of the trained algorithm. This figure illustrates a notable trend: collisions are frequent in the early training episodes but gradually diminish to zero as training progresses.\par

\begin{figure}[ht!]
    \centering
    \begin{subfigure}[c]{0.48\textwidth}
        \includegraphics[width=\textwidth]{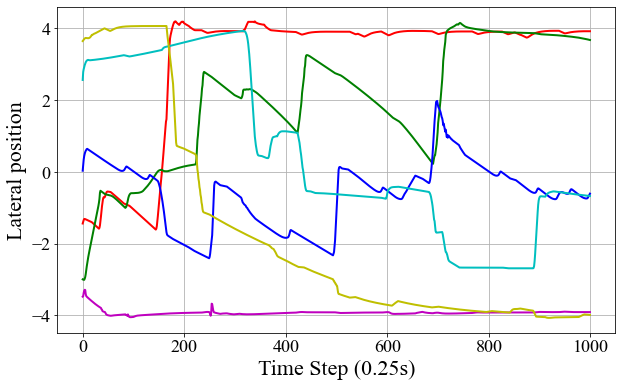}
        \caption{}
        \label{lat.Pos}
    \end{subfigure}
    \begin{subfigure}[c]{0.48\textwidth}
        \includegraphics[width=\textwidth]{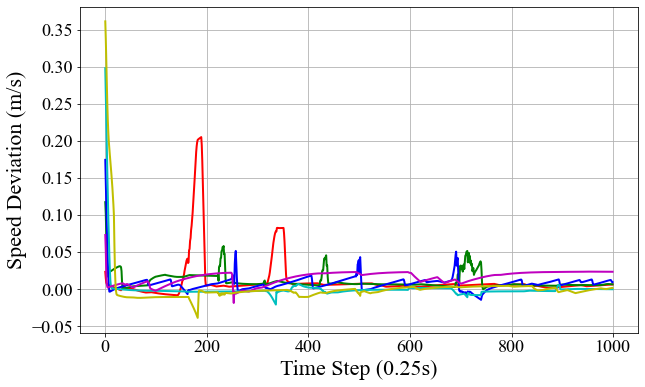}
        \caption{}
        \label{speedDev}
    \end{subfigure}
    \begin{subfigure}[c]{0.48\textwidth}
        \includegraphics[width=\textwidth]{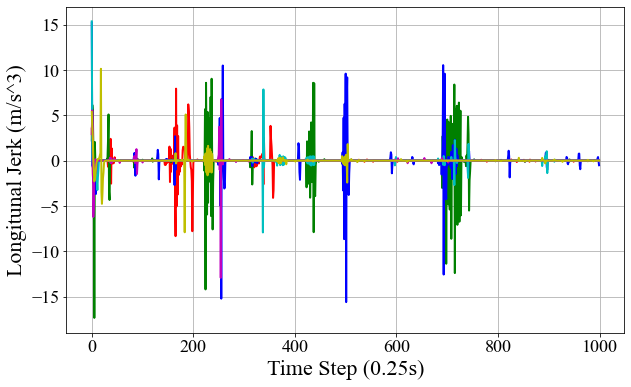}
        \caption{}
        \label{long.Jerk}
    \end{subfigure}
        \begin{subfigure}[c]{0.48\textwidth}
        \includegraphics[width=\textwidth]{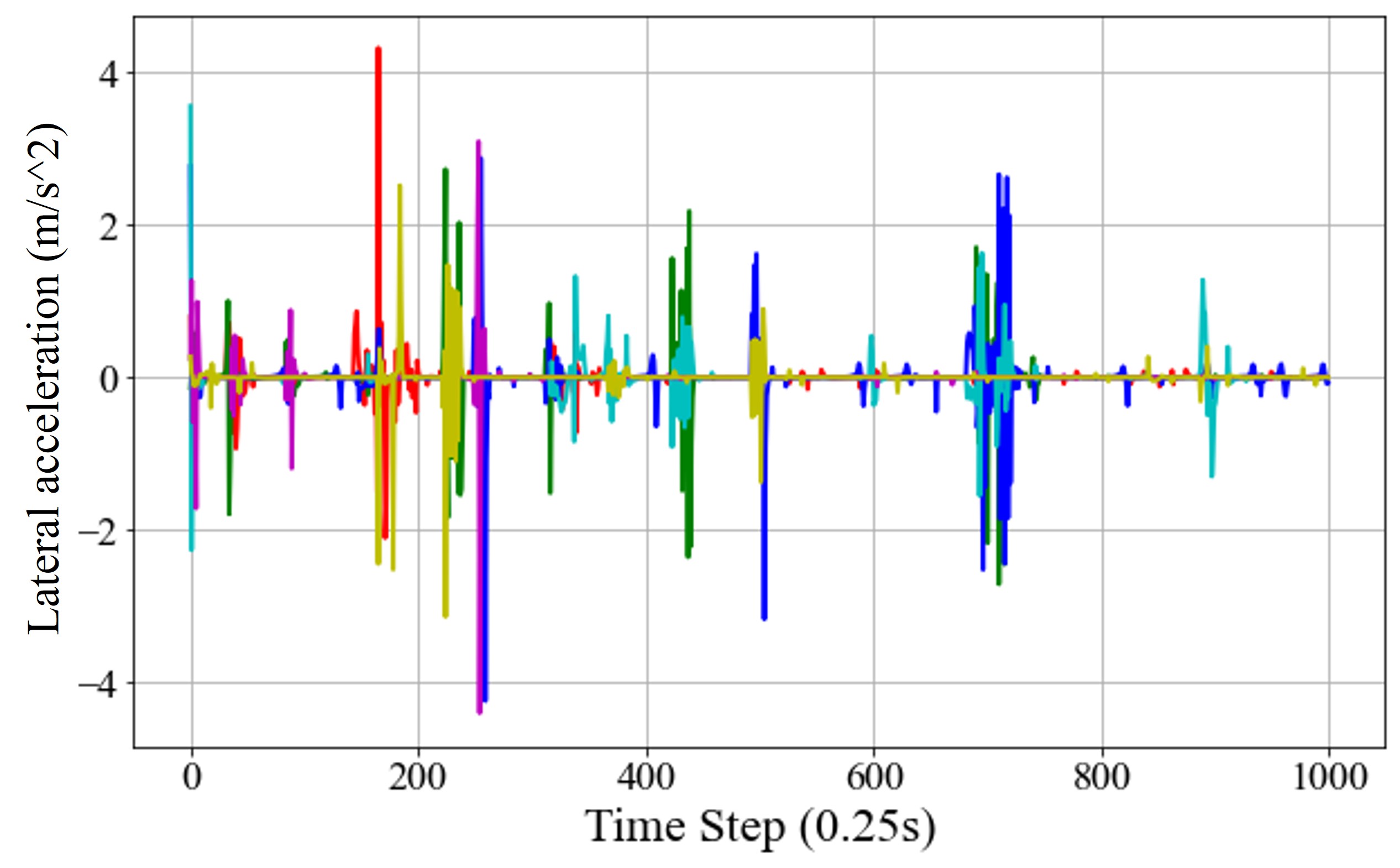}
        \caption{}
        \label{lat.Jerk}
    \end{subfigure}
            %\centering
            \begin{subfigure}{1\textwidth}
            \centering
        \includegraphics[width=.8\textwidth]{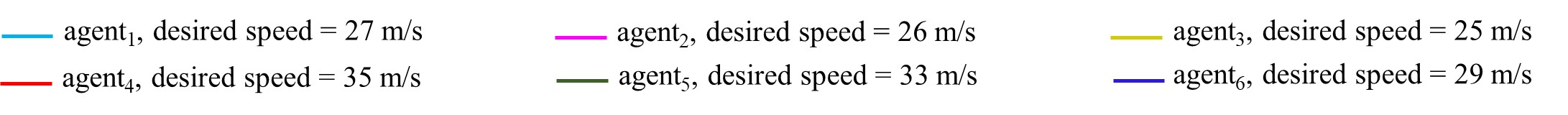}      
    \end{subfigure}
    \caption{Driving behavior agents equipped with the trained MADDPG algorithm on the ring road; lateral maneuvers of agents on the mentioned lane-free ring-road with respect to their following and leading agents (a), speed deviations from the corresponding desired speed (b), longitudinal jerks (c), lateral accelerations (d)}
\label{speedDev_curve}
\end{figure}

To evaluate the performance of the developed MADDPG algorithm, the driving behavior of each of the six agents equipped with the trained MADDPG algorithm on the ring road is evaluated concerning several aspects:  lateral position reflecting the nudging and repulsive forces imposed on them; speed deviation from their corresponding desired speed; longitudinal jerk; and lateral speed variations. Fig.~\ref{speedDev_curve} demonstrates the mentioned criteria as agents drive along the ring road.

For clarity, in SUMO, the lateral position of a vehicle within each lane is represented as a positive value when the vehicle is on the left side of the road and a negative value when the vehicle is on the right side, having a zero value at the center line of the lane. Note that in this work, merely one wide lane is utilized to provide a lane-free traffic environment, and the lateral position of each agent within this lane is shown in Fig.~\ref{lat.Pos}. Since the agents have different desired speeds, they overtake frequently, and thus, their lateral locations change over time as the effect of nudging and repulsive forces, specifically at initial simulation time. Later, they will approach a more stable lateral location depending on their desired speed. For example, although agent\textsubscript{3} is initially located on the left side of the road, it moves to the right side during the simulation due to several interactions with other agents. Conversely, agent\textsubscript{4} and agent\textsubscript{5}, with the highest desired speeds, change their lateral position by moving to the left side through the applied repulsive forces. Therefore, agents are sorted laterally on the road based on their desired speeds, regardless of their initial lateral positions. This behavior facilitates the laminar flow of lane-free traffic, as agents with lower speeds create adequate space on their left side for other vehicles driving at higher velocities.\par
In Fig.~\ref{speedDev}, the speed deviation of each agent from their desired speed is illustrated. As depicted in this figure, within the initial 50 time steps, the mentioned speed deviations of all agents decrease to under 0.01 $m/s$, showcasing the ability of the developed DRL-based algorithm to fulfill the objective of driving next to the desired speed. For agents at certain time steps, such as agent\textsubscript{4} at time step 120 or agent\textsubscript{5} at time step 230, when they initiate an overtaking maneuver, the repulsion force exerted by a slower vehicle in the front causes a temporary decrease in their actual speed, resulting in a positive speed deviation. This deviation is subsequently rectified as the overtaking is completed, and the speed returns to normal. As explained earlier, this repulsion tends to shift vehicles to the left side of the road, as illustrated for the mentioned agents at corresponding time steps in Fig.\ref{lat.Pos}. Conversely, the slow-moving vehicles experience a nudging force, causing a slight acceleration while being overtaken, and they shift slightly to the right. This behavior is exemplified by agent\textsubscript{1} and agent\textsubscript{3}. Consequently, each agent can successfully execute their overtaking actions, maintaining their desired speed as they navigate through the lane-free traffic environment. \par

Figs.~\ref{long.Jerk} and \ref{lat.Jerk} depict the comfort variables, i.e., longitudinal jerks and lateral accelerations of all agents, during their maneuvers on the ring road. Initially, we observe high longitudinal jerk and lateral acceleration values. The reason is that vehicles start with random initial speeds and locations, and thus, they should accelerate or decelerate to match their desired speed and move laterally to avoid collisions.
As the simulation progresses, the value of longitudinal jerks decreases, given that agents optimize their lateral positions concerning their desired speeds. Consequently, the need for drastic lateral and longitudinal actions diminishes, demonstrating the efficiency of the trained algorithm. It is worth noting that, in accordance with (\ref{eq_17}, \ref{eq_19}), and considering the possible drastic actions that each agent can take — longitudinal acceleration of ±4 $m/s^2$, and lateral speed of ±1.5 $m/s$ — the longitudinal jerk at each simulation time step could change between ±32 $m/s^3$, and the lateral acceleration could vary within the range of ±12 $m/s^2$.Given that, we observe that even the sharp changes in Figs.~\ref{long.Jerk} and \ref{lat.Jerk} are well below the maximum of these variations, indicating the comfortable behavior of vehicles. \par
Analyzing the variations in longitudinal and lateral jerk in Fig.~\ref{long.Jerk} and \ref{lat.Jerk}, indicates that the agents with middle desired speeds (e.g., agent\textsubscript{1}, agent\textsubscript{5}, and agent\textsubscript{6}) exhibit the highest lateral and longitudinal jerk. This suggests that these agents engage in more lateral maneuvers, as corroborated by the data in Fig.~\ref{lat.Pos}. Middle-speed agents exhibit increased longitudinal and lateral jerk variations because they consider the states of their rear counterparts, factoring in the imposed nudging force when determining their actions.

\subsection{Evaluation of the test phase}
As mentioned in the previous section, the agents are trained in a multi-agent framework within a ring road with acceptable behavior in terms of defined objectives, including maintaining desired speed, collision-free driving, and adhering to the comfort metrics. It is worth noting that during the training phase, all agents experience identical conditions and are trained in a competitive mode, solely aiming to maximize their individual cumulative reward. Consequently, the behavior of each trained agent within this multi-agent system in the executive mode, in the same state condition, is identical. In other words, each trained agent is interchangeable with others. This implies that we can assign any member of this multi-agent system to control the actions of any vehicles in traffic networks. This happens since, in the executive mode, each agent's actions are solely determined based on its own state, independent of other agents' states. In this section, we use the trained agents and implement them in two test scenarios for further evaluation.

\subsubsection{Ring road scenario}
\begin{figure}[h]
    \centering
    \begin{subfigure}{0.48\textwidth}
        \includegraphics[width=\textwidth]{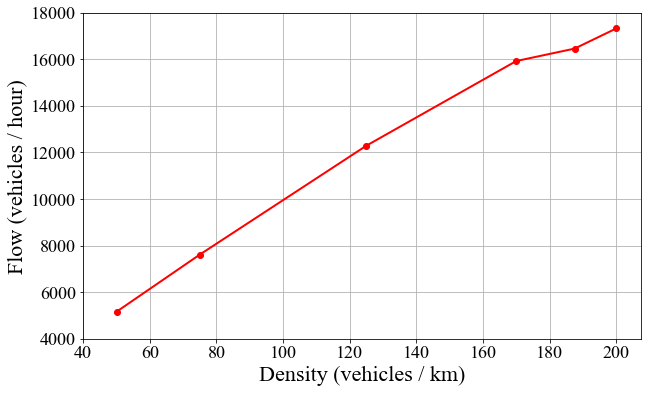}
        \caption{Flow-density diagram}
        \label{fig:FD-flow}
    \end{subfigure}
    \begin{subfigure}{0.48\textwidth}
        \includegraphics[width=\textwidth]{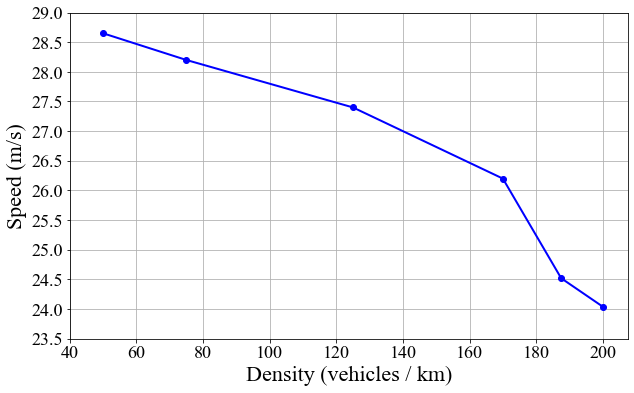}
        \caption{Speed-density diagram}
        \label{fig:FD-speed}
    \end{subfigure}
    \caption{Empirical fundamental diagrams of traffic in the ring road scenario}
    \label{fig:FD}
\end{figure}

In this section, we assumed the same ring road used for training agents and evaluated a higher number of vehicles to test the agents' performance. More specifically, several scenarios with different numbers of vehicles are simulated. In each scenario, the average speed of vehicles and the achieved flow are measured. The obtained empirical fundamental diagrams are then calculated and shown in Fig. \ref{fig:FD}. As expected, the highest speed occurs at the low densities. As the number of vehicles (density) increases, the flow increases as well, but this results in a gradual speed reduction. The highest flow observed is around 17000 $veh/h$ at the density of 200 $veh/km$. This capacity is more than twice of a comparable lane-based road.

\begin{figure}[h]
    \centering
    \includegraphics[width=0.5\textwidth]{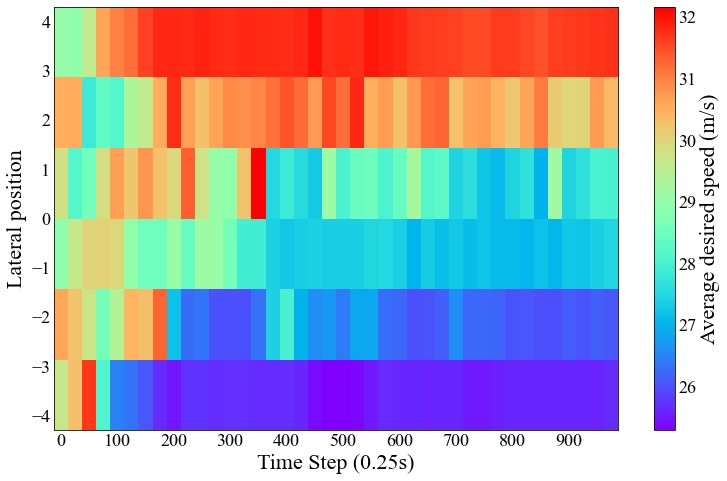}
    \caption{Spatiotemporal distribution of vehicles in the ring road based on their desired speed}
    \label{fig:speed-ringroad}
\end{figure}

As mentioned earlier, we employed the nudging and repulsive forces in such a way as to laterally sort vehicles based on their desired speed. This is illustrated in Fig. \ref{fig:speed-ringroad}. The width of the ring road is spatially discretized into six regions at which the average desired speeds of vehicles are calculated at each time step,

Figure \ref{fig:speed-ringroad} indicates the random initial lateral distribution of vehicles. As time passes, the vehicles are well sorted based on their desired speed, resulting from nudging and repulsion forces. This speed-dependent lateral location of vehicles facilitated overtaking and reduced unnecessary lateral movements and slalom behavior.
\subsubsection{Freeway scenario}
 To scrutinize the performance of the developed algorithm and test the driving behavior of the agents more precisely, a 4 $km$ lane-free road with one off-ramp at the location of 1700 $m$ and one on-ramp at the location of 1800 $m$ is developed. This evaluation focuses on the algorithm's capability to distribute vehicles on a lane-free freeway according to their desired speed and perform merging and diverging maneuvers. Vehicles are initially inserted randomly across the width of the road with initial speeds ranging from 25 $m/s$ to 35 $m/s$. Compared to the ring road scenario, due to SUMO limitations, a very high inflow of vehicles is not possible. Therefore, a medium, still very high compared to the similar lane-based network, inflow rate of 7200 $vehs/hr$ is considered. 
 
\begin{figure}[h]
    \centering
    \begin{subfigure}{0.48\textwidth}
        \includegraphics[width=\textwidth]{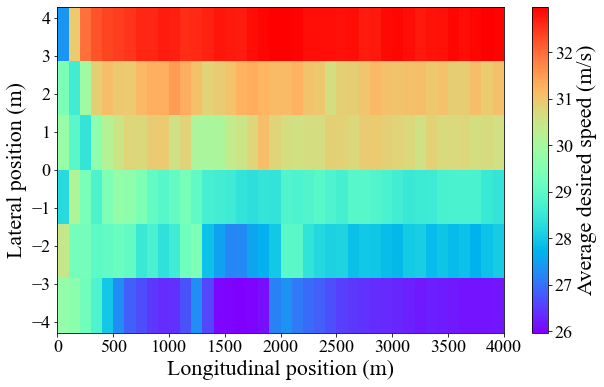}
        \caption{Distribution of the desired speed}
        \label{fig:highway-desired}
    \end{subfigure}
    \begin{subfigure}{0.48\textwidth}
        \includegraphics[width=\textwidth]{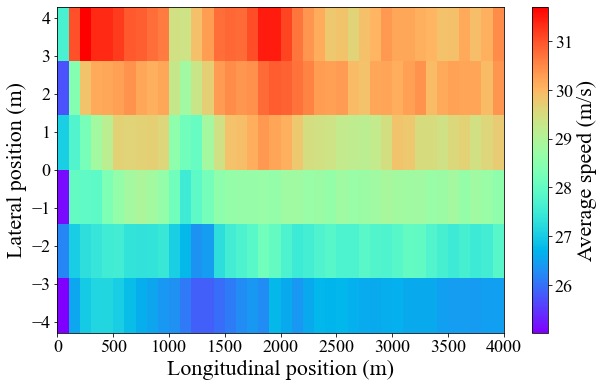}
        \caption{Distribution of the actual speed}
        \label{fig:highway-actual}
    \end{subfigure}
    \caption{Lateral distribution of vehicles in the freeway scenario}
    \label{fig:highway-speed}
\end{figure}

The freeway is segmented spatially, with each section spanning 100 meters in length and possessing a width equivalent to one-sixth of the total road width. The average desired speeds and actual speeds of vehicles traversing each segment are calculated and depicted in Fig.~\ref{fig:highway-speed}. Analysis of both speed distributions indicates that vehicles exhibit lower speeds in the initial part of the freeway, and there is no noticeable correlation between lateral location and speed. This is an expected phenomenon as we randomly insert the vehicles into the road. However, as vehicles progress along the freeway and engage with other vehicles, they tend to become sorted based on their speeds. This sorting circumstance is the direct consequence of the defined nudging and repulsion forces exerted between vehicles.
Fig. \ref{fig:highway-desired} underscores the effectiveness of this sorting mechanism, where vehicles attain lateral positions based on their desired speeds. However, the actual speeds deviate from the desired speeds. Vehicles on the right side of the road tend to exceed their desired speeds, primarily influenced by the nudging effect. Conversely, fast-moving vehicles on the left side typically operate below their desired speeds due to interactions with other vehicles amidst high traffic density and associated repulsive forces. This disparity becomes evident in Fig. \ref{fig:highway-actual}, particularly beyond the on-ramp location. At points of off- and on-ramp connections, where vehicles exiting or entering may not necessarily possess low desired speeds characteristic of the right side of the road, both figures depict a higher speed compared to surrounding areas. Additionally, the merging process initiates a traffic shock wave that propagates laterally across the road, affecting the left side as well.
 
\begin{figure}[h]
\centering
\includegraphics[width=0.85\textwidth]{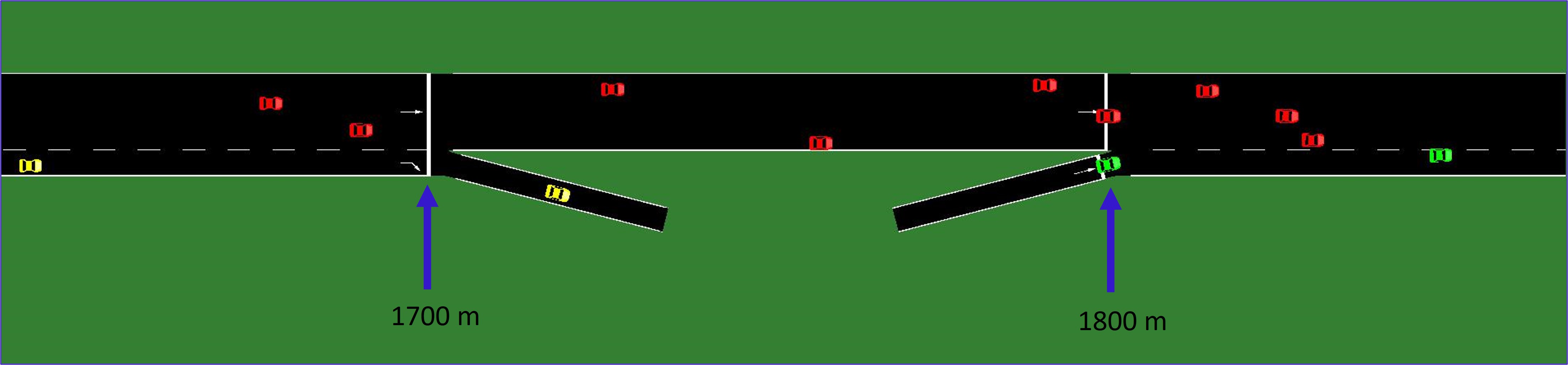}\\
\caption{Merging and diverging behaviors in lane-free traffic environment by means of deceleration and acceleration lanes.}
\label{off/On_ramp}
\end{figure}

As mentioned earlier, one of the imperative objectives of this work is to provide an optimal action policy, which is also utilized for performing merge and diverge maneuvers for each agent on its designated driving route. Fig.~\ref{off/On_ramp} illustrates the traffic network configuration with on/off-ramp. 
These configurations involve the use of extra lanes \textemdash \ acceleration and deceleration lanes \textemdash \ to facilitate merging and diverging maneuvers. In the simulation, the vehicles on the mainstream are shown in red, while yellow and green colors are used to differentiate the vehicles leaving at the off-ramp and entering from the on-ramp, respectively.\par

\begin{figure}[h]
    \centering
    \begin{subfigure}[c]{0.48\textwidth}
        \includegraphics[width=\textwidth]{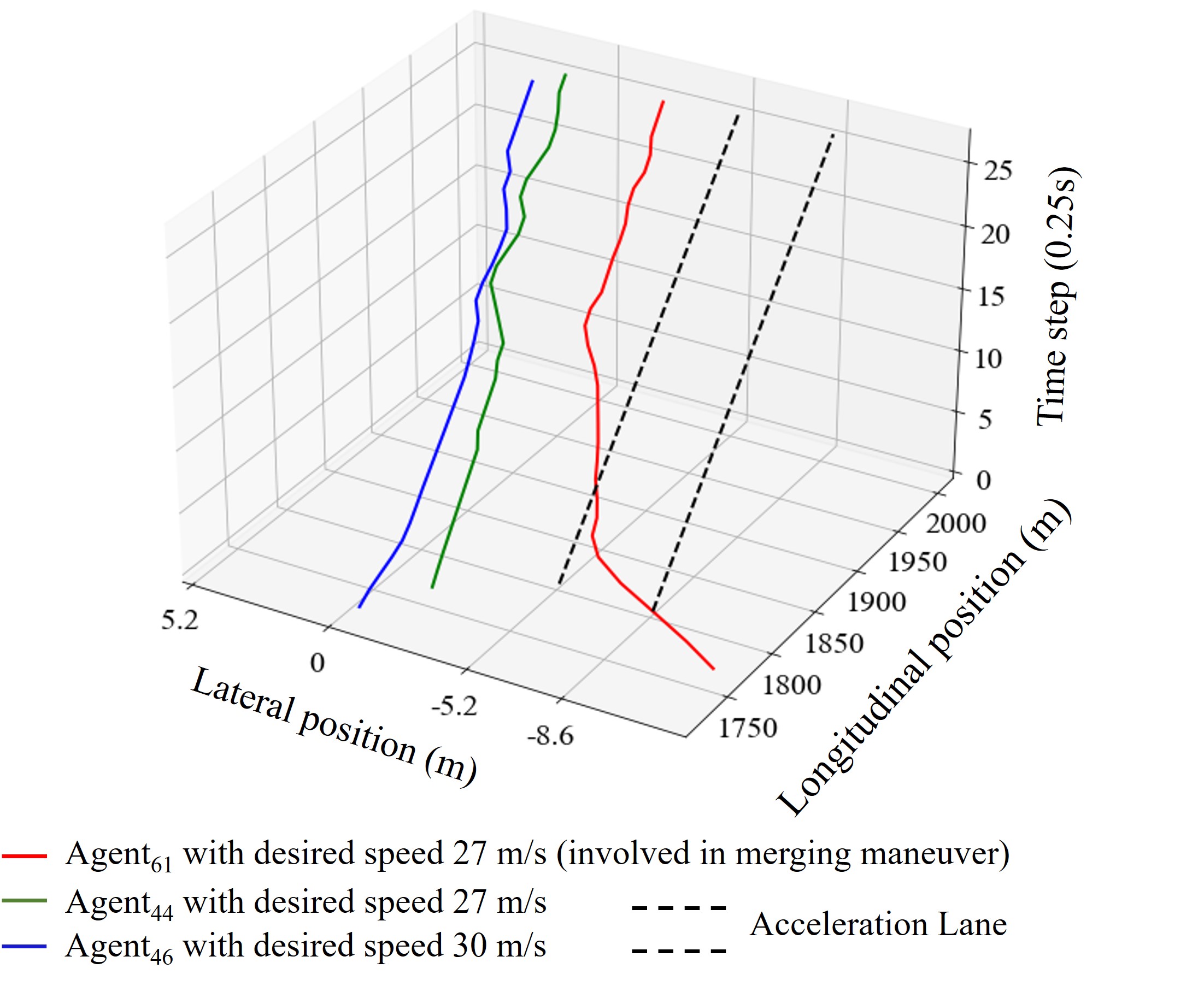}
        \caption{Merging into the mainstream form an on-ramp}
        \label{on-ramp merging}
    \end{subfigure}
    \begin{subfigure}[c]{0.48\textwidth}
        \includegraphics[width=\textwidth]{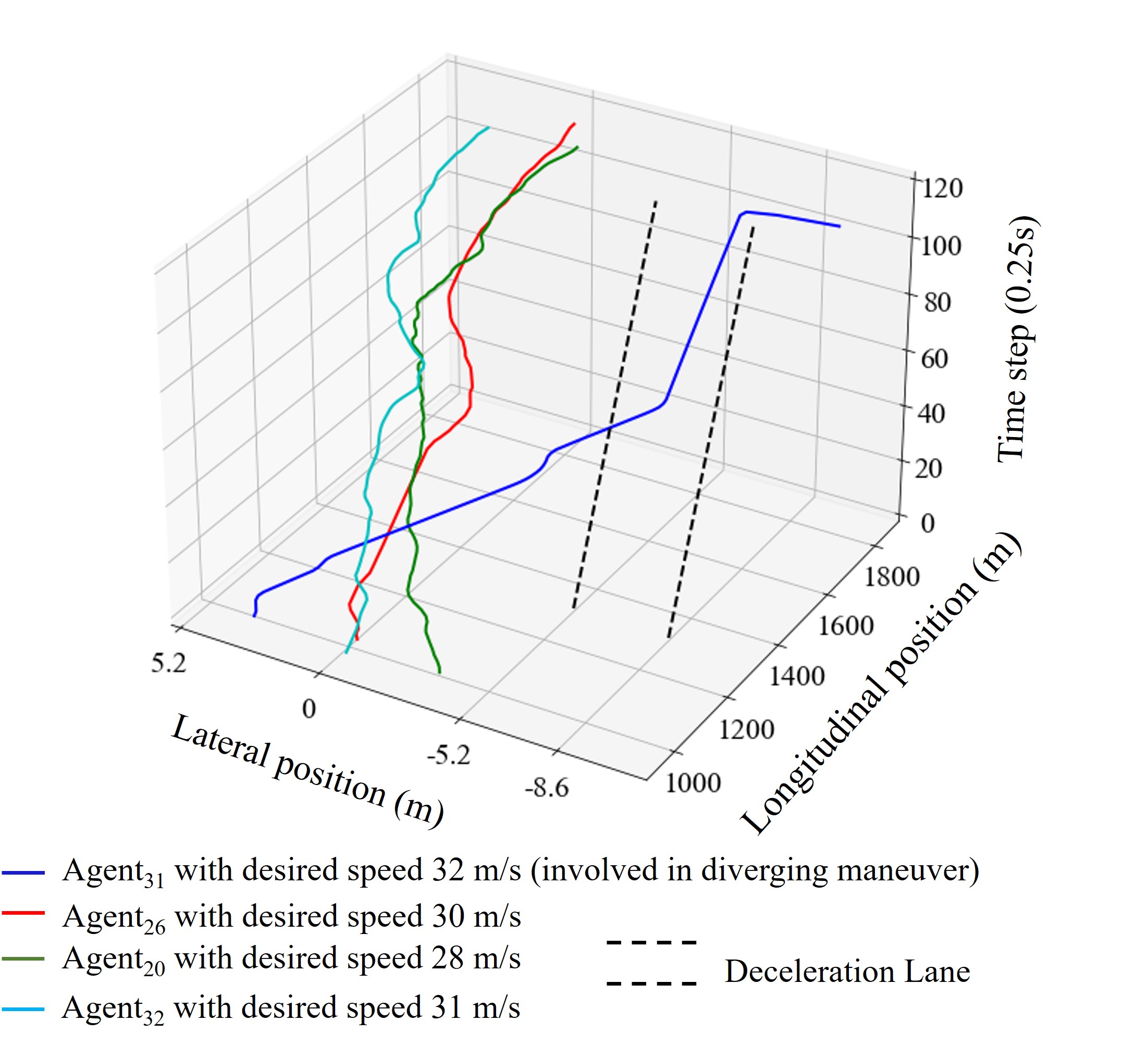}
        \caption{Leaving the mainstream at an off-ramp}
        \label{off-ramp merging}
    \end{subfigure}
 \caption{Trajectories of agents involved in merge and diverge maneuvers and affected agents}
\label{on/off RampCurve}
\end{figure}

%%\begin{figure}[ht!]
 %%   \centering
   %% \begin{subfigure}{0.45\textwidth}
     %%   \includegraphics[width=\textwidth, height=2.7cm]{on-ramp_0}
       %% \caption{Use of an acceleration lane to implement merging at %%an on-ramp.}
        %%\label{on-ramp merging}
    %%\end{subfigure}
    %%\begin{subfigure}{0.45\textwidth}
        %%\includegraphics[width=\textwidth, height=2.7cm]{off-ramp_30}
        %%\caption{Use of a deceleration lane to perform diverging at %%an off-ramp.}
       %% \label{off-ramp merging}
   %% \end{subfigure}
 %%\caption{Merging and diverging behaviors in lane-free traffic environment.}
%%\label{on/off RampMer}
%%\end{figure}
Several agents that are directly or indirectly influenced by such maneuvers are selected to demonstrate and analyze their behavior. In Fig.~\ref{on/off RampCurve}, trajectories of the selected agents are illustrated. Note that in these figures, the initial time of the maneuver is considered as zero. In Fig. \ref{on-ramp merging}, the trajectories of agent\textsubscript{61} during a merging maneuver from the on-ramp together with agent\textsubscript{44} and agent\textsubscript{46} driving on the main road are shown. As depicted, agent\textsubscript{61} after entering the acceleration lane, in the position of 1800 $m$ at time step 3, moves left to merge, prompting agent\textsubscript{44} and agent\textsubscript{46} to also move left to maintain their lateral distance with agent\textsubscript{61} and provide the necessary space for it to merge. As mentioned before, eliptical border intrusion of agent\textsubscript{44} and agent\textsubscript{46} via agent\textsubscript{61} has no effect on agent\textsubscript{61} due to the negligence of nudging force on this agent during the implementation of on-ramp merging. After the completion of the merging maneuver of agent\textsubscript{61}, all agents return to their normal driving behavior on the freeway.\par
In Fig.~\ref{off-ramp merging}, the trajectory of agent\textsubscript{31} leaving at the off-ramp in the presence of three other agents, i.e., agent\textsubscript{20}, agent\textsubscript{26}, and agent\textsubscript{32}, driving on the freeway is illustrated. As depicted earlier, after the initial simulation steps, all agents position themselves laterally based on their desired speed—agents with higher desired speed drive on the left side of the road and vice versa. Once agent\textsubscript{31} initiates the leaving maneuver, at the longitudinal position of 1000 $m$, it gradually shifts to the right side till it enters the deceleration lane at time step 40 in the longitudinal position of 1440 $m$. Finally, agent\textsubscript{31} exits the deceleration lane at time step 98 in the position of 1720 $m$. It is worth mentioning that agent\textsubscript{31} conducts diverging maneuvers before reaching the deceleration lane, as mentioned earlier, to have enough time and longitudinal distance for entering the deceleration in a proper time and with proper speed. The maneuver affects the behavior of other agents as shown in Fig. \ref{off-ramp merging}lane in case of congested traffic.
\section{Conclusion}\label{sec_6}
This paper introduces a novel multi-task lane-free driving strategy that employs a multi-objective multi-agent DRL approach. The proposed MADRL algorithm introduces competitive interactions between agents, fostering a dynamic and non-stationary scenario that closely approximates the complexities of real-world traffic conditions, especially in lane-free traffic networks. Furthermore, we define a specific type of dynamic elliptical borders to create nudging and repulsive forces, playing a crucial role in overtaking, merging, and diverging maneuvers and collision avoidance behavior of the CAVs in this work.\par
After training the mentioned DRL algorithm on a ring road, we evaluated its performance across different networks within various scenarios, including a ring road and a 4 $km$ freeway. The empirical fundamental diagrams obtained from implementing the trained algorithm on a ring road network reveal that a maximum flow of 17000 $veh/h$ is achieved at a density of 200 $veh/km$; this is more than twice the capacity in a lane-based counterpart. \par
Furthermore, with the speed contour plots, we demonstrated that the proposed algorithm has the ability to laterally sort vehicles based on their desired speed. This specific lateral distribution not only facilitates overtaking but also minimizes unnecessary lateral movements and slalom behavior of CAVs in a lane-free traffic network. Additionally, the introduction of a custom lateral reaction of agents to nudging and repulsive forces allows the trained DRL algorithm to effectively handle CAVs' merging and diverging maneuvers.\par
However, given that this MADRL algorithm was trained in relatively low-density traffic conditions, its performance in achieving the predefined objectives diminishes in highly congested traffic. This drawback could be mitigated by fine-tuning the agents of the trained algorithm through a transfer learning approach in a more congested traffic situation, a step we intend to take in our future work.\par
It should also be noted that agents implemented in the current MADDPG algorithm operate with a sole focus on their individual objectives, disregarding the considerations of fellow agents’ objectives. To further enhance the efficacy of our approach, we intend to explore a cooperative multi-agent DRL strategy in future research. This forthcoming investigation aims to develop an approach where agents collaborate to balance their objectives and strategically distribute vehicles across the width of a lane-free freeway. The cooperative multi-agent DRL approach seeks to harmonize the individual goals of agents, with a specific emphasis on maintaining speeds aligned with each vehicle's desired speed. Our future research will involve the development of this cooperative approach, including the clustering of vehicles driving in proximity. These clusters will be treated as distinct cooperative multi-agent DRL groups, fostering collaborative behavior among vehicles on the road.\par
\section*{CRediT Author Statement}

\textbf{Mehran Berahman:} Conceptualization, Methodology, Software, Data Curation, Writing - Original Draft, Visualization.  
\textbf{Majid Rostami-Shahrbabaki:} Conceptualization, Methodology, Software, Data Curation, Writing - Original Draft, Visualization.  
\textbf{Klaus Bogenberger:} Writing - Review \& Editing, Supervision, Project administration.
\section*{Funding} This work at the Technical University of Munich is based on the project “Simulation and
organization of future lane-free traffic”, funded by the German research foundation (DFG) under the
project number BO 5959/1-1.
\section*{Declaration of competing interest} The authors declare that they have no known competing financial interests or personal relationships that could have influenced, or be perceived to influence, the work presented in this paper.\cite{shladover2018connected}

%%\bibliographystyle{elsarticle-num} 
%%\bibliography{LIBRARY_EAAI.bib} 

\end{document}